\documentclass[11pt]{article}

\usepackage[final]{acl}

\usepackage{times}
\usepackage{latexsym}

\usepackage[T1]{fontenc}

\usepackage[utf8]{inputenc}

\usepackage{microtype}

\usepackage{inconsolata}

\usepackage{graphicx}

\usepackage{amsmath}
\usepackage{booktabs}
\usepackage{multirow}
\usepackage{soul}
\usepackage{xcolor}

\definecolor{DrawGray}{HTML}{EEEEEE}
\definecolor{DrawYellow}{HTML}{FFFF88}
\definecolor{DrawCorrelation}{HTML}{FAD7AC}
\definecolor{DrawMetaCorrelation}{HTML}{FAE5C7}

\usepackage[dvipsnames]{xcolor}
\usepackage{tikz}
\usepackage{pgf}
\usepackage{colortbl}

\usepackage{listings}
\newcommand{\gradientcell}[6]{%
    \def\value{#1}%
    \def\minvalue{#2}%
    \def\maxvalue{#3}%
    \def\mincolor{#4}%
    \def\maxcolor{#5}%
    \def\transparency{#6}%
    \def\formattedvalue{\ifdimcomp{\value pt}{<}{0 pt}{}{\hphantom{-}}\value}%
    \ifdimcomp{\value pt}{>}{\maxvalue pt}{\cellcolor{#5!100.0!#4!#6}\formattedvalue}{%
    \ifdimcomp{\value pt}{<}{\minvalue pt}{\cellcolor{#5!0.0!#4!#6}\formattedvalue}{%
         \pgfmathparse{int(round(100*(#1/(\maxvalue-\minvalue))-(\minvalue *(100/(\maxvalue-\minvalue)))))}%
        \xdef\tempa{\pgfmathresult}%
        \cellcolor{#5!\tempa!#4!#6}\formattedvalue%
    }}%
}

\definecolor{LightGray}{gray}{0.95}
\newcommand{\R}[2]{%
    \ifdimcomp{#2 pt}{<}{0.01 pt}{\color{black}}{%
        \ifdimcomp{#2 pt}{>}{0.05 pt}{\color{black!40}}{\color{black!70}}%
    }%
    \gradientcell{#1}{0.0}{0.9}{LightGray}{Goldenrod}{60}%
}
\newcommand{\Cor}[1]{%
    \ifdimcomp{#1 pt}{<}{0 pt}{%
        \gradientcell{#1}{-1}{0}{Blue}{LightGray}{60}%
    }{%
        \gradientcell{#1}{0}{1}{LightGray}{Red}{60}%
    }%
}

\renewcommand{\texttt}[1]{%
  \begingroup
  \ttfamily
  \begingroup\lccode`~=`/\lowercase{\endgroup\def~}{/\discretionary{}{}{}}%
  \begingroup\lccode`~=`[\lowercase{\endgroup\def~}{[\discretionary{}{}{}}%
  \begingroup\lccode`~=`.\lowercase{\endgroup\def~}{.\discretionary{}{}{}}%
  \catcode`/=\active\catcode`[=\active\catcode`.=\active
  \scantokens{#1\noexpand}%
  \endgroup
}

\def\JL#1{\relax}
\def\DH#1{\relax}
\def\LE#1{\relax}
\title{LLM as a Meta-Judge:\\ Synthetic Data for NLP Evaluation Metric Validation}

\author{Lukáš Eigler\textsuperscript{1,2} \and Jindřich Libovický\textsuperscript{1} \and David Hurych\textsuperscript{2} \\
\textsuperscript{1}Faculty of Mathematics and Physics, Charles University, Czech Republic \\
\textsuperscript{2}valeo.ai \\
\texttt{eiglerlukas@gmail.com} \quad \texttt{libovicky@ufal.mff.cuni.cz} \quad   \texttt{david.hurych@gmail.com} 
}

\begin{document}
\maketitle
\begin{abstract}

Validating evaluation metrics for NLG typically relies on expensive and time-consuming human annotations, which predominantly exist only for English datasets. We propose \textit{LLM as a Meta-Judge}, a scalable framework that utilizes LLMs to generate synthetic evaluation datasets via controlled semantic degradation of real data, replacing human judgment. We validate our approach using \textit{meta-correlation}, measuring the alignment between metric rankings derived from synthetic data and those from standard human benchmarks. Experiments across Machine Translation, Question Answering, and Summarization demonstrate that synthetic validation serves as a reliable proxy for human judgment, achieving meta-correlations exceeding 0.9 in multilingual QA, and is a viable alternative when human judgments are unavailable or too expensive to obtain. Our code and data are publicly available at \url{https://github.com/eiglerl/meta-judge}.

\end{abstract}

\section{Introduction}

%

\begin{figure}[t]
\centering\includegraphics[width=.98\columnwidth]{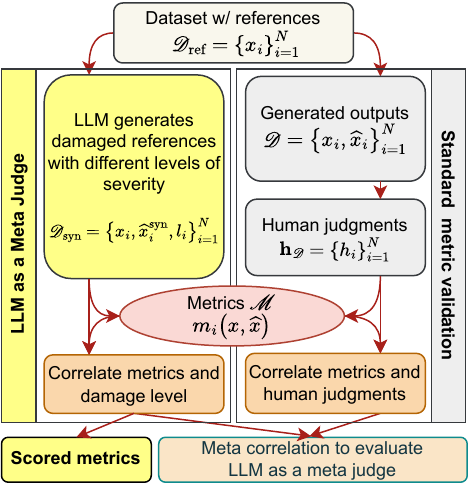}

\caption{\setlength{\fboxsep}{1pt}\colorbox{DrawYellow}{LLM as a Meta-Judge} contrasted with standard \colorbox{DrawGray}{metric validation with human judgment}: LLMs generate damaged reference sentences, and we validate the NLG metrics by \colorbox{DrawCorrelation}{correlation of the damage level} \colorbox{DrawCorrelation}{with metric values}.
We validate our protocol via \colorbox{DrawMetaCorrelation}{meta-correlation}, i.e., correlation with the standard metric validation.}\label{fig:scheme}
\end{figure}

Evaluating natural language generation (NLG) is challenging because semantically equivalent content may have many valid surface realizations. The standard approach to validating evaluation metrics requires collecting human judgments on system outputs and computing correlation with metric scores \citep{graham-etal-2013-continuous}. This creates a bottleneck as human-annotated datasets are expensive, predominantly English-only, and require renewal as systems evolve.

Few public datasets with human judgment exist: WMT for translation \citep{callison-burch-etal-2008-meta}, RoSE for summarization \citep{liu-etal-2023-revisiting}, and MOCHA for question answering \citep{chen-etal-2020-mocha}. These remain largely English-centric, and metrics validated on one task or language do not necessarily transfer to other tasks or languages.

We propose replacing human judgment with LLM-generated outputs of controlled quality. Our approach prompts an LLM to produce semantically degraded versions of reference texts at specified severity levels, creating synthetic data where quality ordering is known by construction (Figure~\ref{fig:scheme}).

We make the following contributions: (1) \emph{Meta-Judge}, a protocol for validating NLG metrics without human judgment, using LLM-generated text with controlled degradation of real reference text as a proxy for system outputs. (2) \emph{meta-correlation}: correlation between metric rankings on synthetic data and on human-annotated benchmarks, measuring proxy reliability. (3) Empirical validation across machine translation, question answering, and summarization in multiple languages, including low-resource languages.

\section{Evaluation Metric Validation}

Evaluation metrics assess generated text quality. We define a metric as a function $m(x, \hat{x})$ computing a real number for reference $x$ and generated string $\hat{x}$. NLG metrics fall into several paradigms.

\emph{String overlap metrics}, such as BLEU \citep{papineni-etal-2002-bleu} and ROUGE \citep{lin-2004-rouge}, compute string-level overlap with interpretability and efficiency, but limited semantic sensitivity.

\emph{Embedding-based metrics} leverage contextual representations: BERTScore \citep{zhang-etal-2020-bertscore} computes token-wise similarity via BERT embeddings, while YiSi \citep{lo-2019-yisi} extends this with semantic role labeling.

\emph{Learned metrics} trained on human judgments achieve the best correlations: COMET \citep{rei-etal-2020-comet} employs cross-lingual encoders, BLEURT \citep{sellam-etal-2020-bleurt} leverages BERT pre-training with synthetic degradation. Recent approaches use LLMs as zero-shot evaluators \citep{zheng-etal-2023-judging,liu-etal-2023-geval}, though this introduces circularity when evaluating LLM-generated text \citep{shen-etal-2023-large}.

Since NLG tasks lack formal specifications, human judgment remains the gold standard. Let $\mathcal{M} = (m_1,\dots,m_k)$ be evaluation metrics, $\mathcal{D} = \{x_i, \hat{x}_i\}^N_{i=1}$ a dataset with references and candidates, and $\mathbf{h}_{\mathcal{D}}$ human judgments. For metric $m_i$, we define scores as a vector $\textbf{s}_{\mathcal{D}}^{m_i} = [m_i(x, \hat{x}): (x,\hat{x}) \in \mathcal{D}]$. Metric quality is measured as correlation with human judgment:
\begin{equation}
r^i_{\text{hum}} = \rho(\textbf{s}_{\mathcal{D}}^{m_i}, \textbf{h}_\mathcal{D})\label{eq:rhum}
\end{equation}
where $\rho$ is Spearman rank correlation.

WMT has systematically evaluated translation metrics since 2008 \citep{callison-burch-etal-2008-meta}, evolving from pairwise comparisons to direct assessment \citep{graham-etal-2013-continuous} and Multidimensional Quality Metrics \citep{freitag-etal-2021-experts}. Datasets exist for summarization (RoSE, 22k annotations; \citealp{liu-etal-2023-revisiting}) and QA (MOCHA, 40k judgments; \citealp{chen-etal-2020-mocha}), though English-only. CUS-QA \citep{libovicky-etal-2025-cus} is a rare multilingual exception. Correlations vary from $r \approx 0.4$ for lexical metrics on summarization to $r > 0.9$ for learned metrics on high-resource translation \citep{freitag-etal-2022-results}.

\section{LLM as a Meta-Judge}

We propose a framework for validating evaluation metrics using an LLM, which we call \textit{Meta-Judge}. Unlike \emph{LLM-as-a-Judge}, where an LLM scores outputs as an evaluation metric, the LLM in the Meta-Judge framework is solely used for data generation to validate evaluation metrics. The protocol is defined as a function $\mathcal{D} \times \mathcal{M} \rightarrow \mathcal{S}$, where we use only reference texts $x$ from dataset $\mathcal{D}$ in this case, $\mathcal{M}$ is a set of evaluation metrics, and $\mathcal{S}$ are metric validation scores.

The protocol consists of three steps: (1) We prompt the LLM to semantically degrade references at known severity levels (damage levels), creating a synthetic dataset with references, damaged references, and pseudo-labels describing the damage. (2) Metrics compute scores for each damaged reference. (3) Correlation between metric scores and pseudo-labels provides metric validation.

Standard metric validation, by contrast, requires candidate outputs and human judgments: $\mathcal{D} \times \mathcal{H} \times \mathcal{M} \rightarrow \mathcal{S}$, where $\mathcal{D}$ contains references and candidates, $\mathcal{H}$ are human judgments, and validation is obtained by correlating metric scores with human judgments.

\subsection{Metric Validation Using Meta-Judge}


The primary application of our approach is to assess metric quality without human judgments. We generate a synthetic dataset $\mathcal{D}_\text{syn}$ using an LLM and an existing dataset. The LLM receives a reference text (and optional context) $x$, along with a specification of damage types through the prompt. We instruct the LLM to generate synthetic text $\hat{x}^\text{syn}$ corresponding to a damage level $l \in \{0, \dots, L_{\max}\}$. These levels serve as pseudo-labels, representing monotonic quality degradation from paraphrasing ($l=0$) to severe hallucinations ($l=L_{\max}$).

The synthetic dataset is $\mathcal{D}_\text{syn} = \{x_i, \hat{x}_i^\text{syn}, l_i\}^N_{i=1}$, where $x_i$ denotes input, optional context, and reference, $\hat{x}_i^\text{syn}$ is the damaged text, and $l_i$ is the damage level. Thus $\hat{x}_i^\text{syn}$ acts as generated output and $l_i$ as pseudo-label replacing human judgment. Collected pseudo-labels are $\textbf{p}_{\mathcal{D}_\text{syn}} = [l: (x,\hat{x}^\text{syn},l) \in \mathcal{D}_\text{syn}]$.

To validate metric $m_i$ on $\mathcal{D}_\text{syn}$, we compute segment-level scores $\textbf{s}^{m_i}_{\mathcal{D}_\text{syn}}$ for all damaged texts. Since metrics measure quality (higher is better) while damage levels measure error (higher is worse), we negate the pseudo-labels:
\begin{equation}
    r^i_\text{syn} = \rho(\textbf{s}^{m_i}_{\mathcal{D}_\text{syn}}, -\textbf{p}_{\mathcal{D}_\text{syn}})\label{eq:rsyn}
\end{equation}

\subsection{Meta-Correlation Analysis} 

For the synthetic dataset to serve as a useful proxy for human judgments, the synthetic correlations $r_\text{syn}$ must accurately estimate the metric's performance. We call this second-order correlation \textit{meta-correlation}.

The validation proceeds in three steps:

(1) Human correlations $r_\text{hum}$. We compute metric scores $\mathbf{s}^{m_i}_{\mathcal{D}}$ for each metric $m_i \in \mathcal{M}$ on dataset $\mathcal{D}$ with human judgments $\mathbf{h}_\mathcal{D}$, then compute Spearman correlation using Equation~\ref{eq:rhum}.

(2) Synthetic correlations $r_\text{syn}$. We compute metric scores $\mathbf{s}^{m_i}_{\mathcal{D}_\text{syn}}$ on the synthetic dataset and correlate with negative pseudo-labels $\mathbf{p}_{\mathcal{D}_\text{syn}}$ using Equation~\ref{eq:rsyn}.

(3) Meta-correlation $\textit{MC}$. We compute Spearman correlation between the vector of human correlations $\mathbf{r}_h = [r_\text{hum}^i: m_i \in \mathcal{M}]$ and synthetic correlations $\mathbf{r}_s = [r_\text{syn}^i: m_i \in \mathcal{M}]$:
\begin{equation}
    \textit{MC} = \rho(\mathbf{r}_h, \mathbf{r}_s)
\end{equation}
A high positive meta-correlation indicates that the synthetic dataset is a reliable proxy for human judgment.

\begin{table*}[ht]
\footnotesize\centering
\setlength{\tabcolsep}{4pt}

\begin{tabular}{l l ccc ccc c c cc ccc}
\toprule
\multirow{2}{*}{Model} & \multirow{2}{*}{Shot} & \multicolumn{3}{c}{CUS-QA (en)} & \multicolumn{3}{c}{CUS-QA (orig.)} & \multirow{2}{*}{\textls[-60]{MOCHA}} & \multirow{2}{*}{RoSE} & \multicolumn{2}{c}{WMT 21} & \multicolumn{3}{c}{WMT 24} \\
\cmidrule(lr){3-5} \cmidrule(lr){6-8} \cmidrule(lr){11-12} \cmidrule(lr){13-15}
 & & cs & sk & uk & cs & sk & uk & & & en-ha & xh-zu & cs-uk & en-cs & en-is \\
\midrule
\multirow{2}{*}{Llama 4 Scout} & Few & \R{.859}{.0000} & \R{.829}{.0000} & \R{.731}{.0000} & \R{.895}{.0000} & \R{.913}{.0000} & \R{.716}{.0000} & \R{.302}{.0594} & \R{.473}{.0055} & \R{.543}{.0014} & \R{.523}{.0021} & \R{.945}{.0000} & \R{.276}{.0777} & \R{.299}{.0608} \\
 & Zero & \R{.793}{.0000} & \R{.774}{.0000} & \R{.669}{.0000} & \R{.827}{.0000} & \R{.875}{.0000} & \R{.717}{.0000} & \R{.211}{.1409} & \R{.453}{.0077} & \R{.303}{.0584} & \R{.473}{.0055} & \R{.920}{.0000} & \R{.472}{.0056} & \R{.413}{.0145} \\
\midrule
\multirow{2}{*}{Llama 3.3 70B} & Few & \R{.808}{.0000} & \R{.788}{.0000} & \R{.751}{.0000} & \R{.917}{.0000} & \R{.922}{.0000} & \R{.759}{.0000} & \R{.678}{.0000} & \R{.043}{.4146} & \R{.490}{.0040} & \R{.454}{.0076} & \R{.854}{.0000} & \R{.410}{.0150} & \R{.419}{.0132} \\
 & Zero & \R{.621}{.0002} & \R{.652}{.0001} & \R{.583}{.0006} & \R{.777}{.0000} & \R{.774}{.0000} & \R{.605}{.0004} & \R{.486}{.0044} & \R{.327}{.0448} & \R{.432}{.0108} & \R{.447}{.0086} & \R{.794}{.0000} & \R{.491}{.0040} & \R{.513}{.0026} \\
\midrule
\multirow{2}{*}{Qwen 3 30B} & Few & \R{.796}{.0000} & \R{.792}{.0000} & \R{.651}{.0001} & \R{.871}{.0000} & \R{.885}{.0000} & \R{.684}{.0000} & \R{.726}{.0000} & \R{.833}{.0000} & \R{.334}{.0412} & \R{.285}{.0707} & \R{.918}{.0000} & \R{.325}{.0457} & \R{.370}{.0263} \\
 & Zero & \R{.917}{.0000} & \R{.776}{.0000} & \R{.754}{.0000} & \R{.956}{.0000} & \R{.955}{.0000} & \R{.821}{.0000} & \R{.872}{.0000} & \R{.675}{.0000} & \R{.032}{.4352} & \R{.308}{.0553} & \R{.937}{.0000} & \R{.286}{.0699} & \R{.269}{.0829} \\
\bottomrule
\end{tabular}

\caption{Meta-correlation between LLM as Meta-Judge and standard metric validation with human judgment measured by Spearman correlation. The values are black if they are significant at the 0.01 confidence, \textcolor{black!70}{dark gray} if they are only significant at 0.05 confidence, and \textcolor{black!40}{light gray} if they are not significantly different from zero.}
\label{tab:spearman}
\end{table*}

\section{Experiments}

\subsection{Tasks}

We evaluate our method across three tasks and multiple languages, including low-resource settings. All selected tasks have datasets with human judgments necessary for computing meta-correlation.

\paragraph{Question Answering.} CUS-QA \citep{libovicky-etal-2025-cus} covers region-specific knowledge in Czech, Slovak, Ukrainian, and their English translations. MOCHA \citep{chen-etal-2020-mocha} tests various reasoning forms compiled from multiple sources, with judgments collected on a 1--5 scale and averaged across annotators.

\paragraph{Summarization.} RoSE \citep{liu-etal-2023-revisiting} provides human judgments on news article summarization via Atomic Content Unit matching, aggregated over three annotators per summary.

\paragraph{Machine Translation.}  We use language pairs from WMT 2021 \citep{akhbardeh-etal-2021-findings} (English (en) to Hausa (ha), Xhosa (xh) to Zulu (zu)), which uses Direct Assessment with z-normalized scores, and WMT 2024 \citep{kocmi-etal-2024-findings} (Czech (cs) to Ukrainian (uk), English (en) to Czech (cs), English (en) to Icelandic (is)), which uses Error Span Annotation, covering both high-resource and low-resource settings.

\subsection{Automatic Metrics}

We use a diverse set of seven evaluation metrics spanning multiple paradigms: string overlap metrics (BLEU, ROUGE, chrF, METEOR), embedding-based metrics (BERTScore), and model-based approaches (COMET, BLEURT). To get a sufficient number of data points for robust correlation estimation, we evaluate each metric under multiple parameter configurations. Details of metric parameters are provided in Appendix~\ref{app:parameters}.

\subsection{Synthetic Data Generation}

To generate synthetic datasets containing semantically damaged texts, we use three open-source LLMs with varying parameter counts and architectures. We prompt the LLMs to produce semantically damaged outputs based on the given input and discrete damage level, utilizing greedy decoding to ensure deterministic generation. We use six damage levels ($L_{\max}=5$, levels $0$--$5$) as a design choice. While any range can be used, the prompt needs to describe every level of damage, and defining many distinct levels is difficult due to potential overlap and nuance. We compare three Meta-Judge models: \textit{Llama 4 Scout} \citep{llama4}, \textit{Llama 3.3 70B} \citep{grattafiori-etal-2024-llama3}, and \textit{Qwen 3 30B} \citep{qwen3}.

To investigate the impact of in-context learning on the consistency of these pseudo-labels, we employ two prompting strategies: 
(1) \emph{Zero-Shot:} The model relies solely on the instruction and damage definitions, and 
(2) \emph{Few-Shot:} The model is provided with domain-specific examples of damage levels to better steer the degradation process. 
The complete prompts for all tasks under both prompting conditions are provided in Appendix~\ref{sec:prompts}. The few-shot examples were manually prepared according to the damage level specifications.

\begin{table}
\centering\footnotesize

\begin{tabular}{ll cc}
\toprule
Model & Shot & Spearman & Kendall \\
\midrule
\multirow{2}{*}{Llama 4 Scout} & Few  & $.905 \pm .010$ & $.756 \pm .020$ \\
                               & Zero & $.884 \pm .056$ & $.710 \pm .087$ \\
\midrule
\multirow{2}{*}{Llama 3.3 70B} & Few  & $.914 \pm .019$ & $.750 \pm .024$ \\
                               & Zero & $.862 \pm .051$ & $.733 \pm .046$ \\
\midrule
\multirow{2}{*}{Qwen 3 30B}    & Few  & $.890 \pm .029$ & $.731 \pm .049$ \\
                               & Zero & $.958 \pm .022$ & $.827 \pm .035$ \\
\bottomrule
\end{tabular}

\caption{Mean $\pm$ standard deviation of the meta correlation for the Czech subset of CUS-QA with several prompts.}
\label{tab:robustness}
\end{table}

\subsection{Results}

Table~\ref{tab:spearman} reports meta-correlation results across all datasets and tasks using Spearman rank correlation (Kendall correlation in Table~\ref{tab:kendall}; detailed per-metric correlations in Tables~\ref{tab:qa_cz_combined_spearman}--\ref{tab:wmt24_enis_spearman} in the Appendix).

The strongest and most consistent results are observed in CUS-QA (question answering). The Meta-Judge protocol achieves high meta-correlation across all tested languages (Czech, Slovak, Ukrainian) and their English translations, with values exceeding 0.9 in several configurations. Performance is generally higher for original languages than translations. Results on MOCHA are similarly strong, with Qwen 3 reaching 0.87 in zero-shot mode, though Llama 4 Scout below other models.

Note that few-shot results are not always better than zero-shot. This was revealed by multiple benchmarks on the internet, but, as far as we know, no scientific work studied this problem thoroughly for our tasks. Similar behavior of Qwen and Llama models was observed for the chain-of-thought prompting by \citet{cheng-etal-2025}.

Meta-correlation results for RoSE (summarization) and WMT (machine translation) are more variable. WMT 2024 Czech--Ukrainian achieves high meta-correlation across all models, but this is partially an artifact: the default ROUGE tokenizer discards Cyrillic characters, causing ROUGE to fail on Ukrainian text and produce clearly poor metrics that make the overall ranking easier. For other language pairs, we attribute variability to differences in system output variance: English--Czech is a long-standing WMT task where systems consistently achieve high performance, making ranking difficult, while lower-resource pairs (Icelandic, Zulu, Hausa) exhibit greater output variance that facilitates metric discrimination. Compared to Llama models, Qwen performs notably worse on low-resource WMT 2021 languages.

Per-metric analysis (Tables~\ref{tab:qa_cz_combined_spearman}--\ref{tab:wmt24_enis_spearman}) reveals consistent patterns: BLEU shows low or negative correlations with both human judgments and synthetic damage levels, with the correlation getting lower as the $n$-gram order increases. In contrast, chrF performs reliably across all configurations, often matching or exceeding learned metrics, suggesting character-level overlap captures semantic degradation more robustly than word-level $n$-grams.

Language-specific analysis shows substantially lower correlations for Ukrainian text in CUS-QA, partially due to tokenization issues. ROUGE-4 produces undefined values for Cyrillic entirely. English translations exhibit lower correlations with human judgment than the original languages, yet synthetic correlations remain stable, indicating robustness to translation-induced noise.

\subsection{Prompt Robustness}
\label{sec:robustness}

To assess sensitivity to the precise wording of damage instructions, we tested five prompt variants per model on the Czech (original) CUS-QA dataset split. Table~\ref{tab:robustness} reports mean and standard deviation of the meta-correlation. Standard deviations remain below 0.06 in all cases with Spearman correlation. This indicates that the Meta-Judge protocol is robust to the exact formulation of the damage descriptions. 

\section{Related Work}

Synthetic data for NLG evaluation has been explored with different objectives. \citet{sellam-etal-2020-bleurt} used random perturbations (mask-filling, backtranslation, word dropout) to generate training data for BLEURT, focusing on metric \emph{training} rather than \emph{validation}. More closely related, \citet{deviyani-diaz-2025-contextual} introduced local metric accuracy using rule-based and LLM-based perturbations to analyze how metric performance varies across evaluation contexts. Our Meta-Judge framework differs by validating whether synthetic degradation can serve as a \emph{universal proxy} for human evaluation across tasks and languages.

The meta-correlation concept has been applied by \citet{shen-etal-2023-large}, who examined how LLM-metric agreement with human judgment degrades for higher-quality outputs. While they assess single-evaluator reliability across quality levels, our meta-correlation validates whether synthetic data preserves the relative ranking of \emph{multiple metrics}, enabling metric validation without human labels.

\section{Conclusions}

We introduced LLM as a Meta-Judge, a protocol for validating NLG evaluation metrics using LLM-generated synthetic data with controlled semantic degradation, eliminating the need for expensive human annotation.
Our meta-correlation analysis demonstrates that synthetic evaluation can serve as a reliable proxy for human judgment, particularly for question-answering tasks, where we achieve meta-correlations exceeding 0.9. The approach proves most effective in high-resource languages and shows promise for low-resource settings where human-annotated evaluation data is scarce or unavailable.
The results vary across tasks, with stronger performance on QA compared to summarization and machine translation. The protocol provides a scalable alternative for metric validation when human evaluation is impractical.

\section*{Limitations}

The reliability of synthetic data generation depends on the LLM's proficiency in the target language. For low-resource languages, the quality of semantic degradations may be inconsistent, as reflected in our lower meta-correlations for Hausa, Zulu, and Xhosa.

Our method requires specifying what types of errors the metrics should detect. The damage definitions in our prompts are task-specific, and applying the framework to new generation tasks requires designing appropriate degradation strategies based on domain knowledge.

A further concern is circularity, where the model used for data generation and the evaluation metric(s) share architecture and/or training data. We consider circularity to be a critical issue primarily when the LLM used in LLM-as-a-Judge matches the generator LLM, which is not the case in our setup.  

Finally, validating the Meta-Judge approach requires datasets with human judgments to compute meta-correlation. For new tasks or languages without existing human-annotated evaluation data, limited pilot annotations may still be necessary to verify the method's reliability.

\section*{Acknowledgement}

This research was supported by the Charles University project PRIMUS/23/SCI/023 and project CZ.02.01.01/00/23\_020/0008518 of the Ministry of Education, Youth and Sports of the Czech Republic. Computational resources were provided by the Ministry of Education, Youth and Sports of the Czech Republic through the e-INFRA CZ (ID:90254).

We used GitHub Copilot to assist in writing the source code for our experiments. We used Claude and Gemini for formatting tables, spell checking, and text shortening.


\bibliography{custom}

@inproceedings{cheng-etal-2025,
  address      = {Suzhou, China},
  author       = {Xiang Cheng and
                  Chengyan Pan and
                  Minjun Zhao and
                  Deyang Li and
                  Fangchao Liu and
                  Xinyu Zhang and
                  Xiao Zhang and
                  Yong Liu},
  bibsource    = {dblp computer science bibliography, https://dblp.org},
  booktitle    = {Findings of the Association for Computational Linguistics: {EMNLP} 2025},
  editor       = {Christos Christodoulopoulos and
                  Tanmoy Chakraborty and
                  Carolyn Rose and
                  Violet Peng},
  pages        = {13533--13554},
  publisher    = {Association for Computational Linguistics},
  title        = {{R}evisiting {C}hain-of-Thought {P}rompting: {Z}ero-shot {C}an {B}e {S}tronger than {F}ew-shot},
  url          = {https://aclanthology.org/2025.findings-emnlp.729/},
  year         = {2025}
}

@inproceedings{papineni-etal-2002-bleu,
  address      = {Philadelphia, Pennsylvania, USA},
  author       = {Papineni, Kishore  and
                  Roukos, Salim  and
                  Ward, Todd  and
                  Zhu, Wei-Jing},
  booktitle    = {Proceedings of the 40th Annual Meeting of the Association for Computational Linguistics},
  doi          = {10.3115/1073083.1073135},
  editor       = {Isabelle, Pierre  and
                  Charniak, Eugene  and
                  Lin, Dekang},
  month        = jul,
  pages        = {311--318},
  publisher    = {Association for Computational Linguistics},
  title        = {{B}leu: a Method for Automatic Evaluation of Machine Translation},
  url          = {https://aclanthology.org/P02-1040/},
  year         = {2002}
}

@inproceedings{rei-etal-2020-comet,
  address      = {Online},
  author       = {Rei, Ricardo  and
                  Stewart, Craig  and
                  Farinha, Ana C  and
                  Lavie, Alon},
  booktitle    = {Proceedings of the 2020 Conference on Empirical Methods in Natural Language Processing (EMNLP)},
  doi          = {10.18653/v1/2020.emnlp-main.213},
  editor       = {Webber, Bonnie  and
                  Cohn, Trevor  and
                  He, Yulan  and
                  Liu, Yang},
  month        = nov,
  pages        = {2685--2702},
  publisher    = {Association for Computational Linguistics},
  title        = {{COMET}: A Neural Framework for {MT} Evaluation},
  url          = {https://aclanthology.org/2020.emnlp-main.213/},
  year         = {2020}
}

@inproceedings{deviyani-diaz-2025-contextual,
  address      = {Albuquerque, New Mexico},
  author       = {Deviyani, Athiya  and
                  Diaz, Fernando},
  booktitle    = {Findings of the Association for Computational Linguistics: NAACL 2025},
  doi          = {10.18653/v1/2025.findings-naacl.276},
  editor       = {Chiruzzo, Luis  and
                  Ritter, Alan  and
                  Wang, Lu},
  isbn         = {979-8-89176-195-7},
  month        = apr,
  pages        = {4906--4925},
  publisher    = {Association for Computational Linguistics},
  title        = {Contextual Metric Meta-Evaluation by Measuring Local Metric Accuracy},
  url          = {https://aclanthology.org/2025.findings-naacl.276/},
  year         = {2025}
}

@inproceedings{shen-etal-2023-large,
  address      = {Singapore},
  author       = {Shen, Chenhui  and
                  Cheng, Liying  and
                  Nguyen, Xuan-Phi  and
                  You, Yang  and
                  Bing, Lidong},
  booktitle    = {Findings of the Association for Computational Linguistics: EMNLP 2023},
  doi          = {10.18653/v1/2023.findings-emnlp.278},
  editor       = {Bouamor, Houda  and
                  Pino, Juan  and
                  Bali, Kalika},
  month        = dec,
  pages        = {4215--4233},
  publisher    = {Association for Computational Linguistics},
  title        = {Large Language Models are Not Yet Human-Level Evaluators for Abstractive Summarization},
  url          = {https://aclanthology.org/2023.findings-emnlp.278/},
  year         = {2023}
}

@inproceedings{lin-2004-rouge,
  address      = {Barcelona, Spain},
  author       = {Lin, Chin-Yew},
  booktitle    = {Text Summarization Branches Out},
  month        = jul,
  pages        = {74--81},
  publisher    = {Association for Computational Linguistics},
  title        = {{ROUGE}: {A} {P}ackage for {A}utomatic {E}valuation of {S}ummaries},
  url          = {https://aclanthology.org/W04-1013/},
  year         = {2004}
}

@inproceedings{zhang-etal-2020-bertscore,
  author       = {Tianyi Zhang and
                  Varsha Kishore and
                  Felix Wu and
                  Kilian Q. Weinberger and
                  Yoav Artzi},
  bibsource    = {dblp computer science bibliography, https://dblp.org},
  booktitle    = {8th International Conference on Learning Representations, {ICLR} 2020,
                  Addis Ababa, Ethiopia, April 26-30, 2020},
  publisher    = {OpenReview.net},
  title        = {BERTScore: Evaluating Text Generation with {BERT}},
  url          = {https://openreview.net/forum?id=SkeHuCVFDr},
  year         = {2020}
}

@inproceedings{lo-2019-yisi,
  address      = {Florence, Italy},
  author       = {Lo, Chi-kiu},
  booktitle    = {Proceedings of the Fourth Conference on Machine Translation (Volume 2: Shared Task Papers, Day 1)},
  doi          = {10.18653/v1/W19-5358},
  editor       = {Bojar, Ond{\v{r}}ej  and
                  Chatterjee, Rajen  and
                  Federmann, Christian  and
                  Fishel, Mark  and
                  Graham, Yvette  and
                  Haddow, Barry  and
                  Huck, Matthias  and
                  Yepes, Antonio Jimeno  and
                  Koehn, Philipp  and
                  Martins, Andr{\'e}  and
                  Monz, Christof  and
                  Negri, Matteo  and
                  N{\'e}v{\'e}ol, Aur{\'e}lie  and
                  Neves, Mariana  and
                  Post, Matt  and
                  Turchi, Marco  and
                  Verspoor, Karin},
  month        = aug,
  pages        = {507--513},
  publisher    = {Association for Computational Linguistics},
  title        = {{Y}i{S}i - a {U}nified {S}emantic {MT} {Q}uality {E}valuation and {E}stimation {M}etric for {L}anguages with {D}ifferent {L}evels of {A}vailable {R}esources},
  url          = {https://aclanthology.org/W19-5358/},
  year         = {2019}
}

@inproceedings{zheng-etal-2023-judging,
  author       = {Lianmin Zheng and
                  Wei{-}Lin Chiang and
                  Ying Sheng and
                  Siyuan Zhuang and
                  Zhanghao Wu and
                  Yonghao Zhuang and
                  Zi Lin and
                  Zhuohan Li and
                  Dacheng Li and
                  Eric P. Xing and
                  Hao Zhang and
                  Joseph E. Gonzalez and
                  Ion Stoica},
  bibsource    = {dblp computer science bibliography, https://dblp.org},
  booktitle    = {Advances in Neural Information Processing Systems 36: Annual Conference
                  on Neural Information Processing Systems 2023, NeurIPS 2023, New Orleans,
                  LA, USA, December 10 - 16, 2023},
  editor       = {Alice Oh and
                  Tristan Naumann and
                  Amir Globerson and
                  Kate Saenko and
                  Moritz Hardt and
                  Sergey Levine},
  title        = {Judging LLM-as-a-Judge with MT-Bench and Chatbot Arena},
  url          = {http://papers.nips.cc/paper\_files/paper/2023/hash/91f18a1287b398d378ef22505bf41832-Abstract-Datasets\_and\_Benchmarks.html},
  year         = {2023}
}

@inproceedings{liu-etal-2023-geval,
  address      = {Singapore},
  author       = {Liu, Yang  and
                  Iter, Dan  and
                  Xu, Yichong  and
                  Wang, Shuohang  and
                  Xu, Ruochen  and
                  Zhu, Chenguang},
  booktitle    = {Proceedings of the 2023 Conference on Empirical Methods in Natural Language Processing},
  doi          = {10.18653/v1/2023.emnlp-main.153},
  editor       = {Bouamor, Houda  and
                  Pino, Juan  and
                  Bali, Kalika},
  month        = dec,
  pages        = {2511--2522},
  publisher    = {Association for Computational Linguistics},
  title        = {{G}-Eval: {NLG} Evaluation using Gpt-4 with Better Human Alignment},
  url          = {https://aclanthology.org/2023.emnlp-main.153/},
  year         = {2023}
}

@inproceedings{callison-burch-etal-2008-meta,
  address      = {Columbus, Ohio},
  author       = {Callison-Burch, Chris  and
                  Fordyce, Cameron  and
                  Koehn, Philipp  and
                  Monz, Christof  and
                  Schroeder, Josh},
  booktitle    = {Proceedings of the Third Workshop on Statistical Machine Translation},
  editor       = {Callison-Burch, Chris  and
                  Koehn, Philipp  and
                  Monz, Christof  and
                  Schroeder, Josh  and
                  Fordyce, Cameron Shaw},
  month        = jun,
  pages        = {70--106},
  publisher    = {Association for Computational Linguistics},
  title        = {Further Meta-Evaluation of Machine Translation},
  url          = {https://aclanthology.org/W08-0309/},
  year         = {2008}
}

@inproceedings{graham-etal-2013-continuous,
  address      = {Sofia, Bulgaria},
  author       = {Graham, Yvette  and
                  Baldwin, Timothy  and
                  Moffat, Alistair  and
                  Zobel, Justin},
  booktitle    = {Proceedings of the 7th Linguistic Annotation Workshop and Interoperability with Discourse},
  editor       = {Pareja-Lora, Antonio  and
                  Liakata, Maria  and
                  Dipper, Stefanie},
  month        = aug,
  pages        = {33--41},
  publisher    = {Association for Computational Linguistics},
  title        = {Continuous Measurement Scales in Human Evaluation of Machine Translation},
  url          = {https://aclanthology.org/W13-2305/},
  year         = {2013}
}

@article{freitag-etal-2021-experts,
  address      = {Cambridge, MA},
  author       = {Freitag, Markus  and
                  Foster, George  and
                  Grangier, David  and
                  Ratnakar, Viresh  and
                  Tan, Qijun  and
                  Macherey, Wolfgang},
  doi          = {10.1162/tacl_a_00437},
  editor       = {Roark, Brian  and
                  Nenkova, Ani},
  journal      = {Transactions of the Association for Computational Linguistics},
  pages        = {1460--1474},
  publisher    = {MIT Press},
  title        = {Experts, Errors, and Context: A Large-Scale Study of Human Evaluation for Machine Translation},
  url          = {https://aclanthology.org/2021.tacl-1.87/},
  volume       = {9},
  year         = {2021}
}

@inproceedings{liu-etal-2023-revisiting,
  address      = {Toronto, Canada},
  author       = {Liu, Yixin  and
                  Fabbri, Alex  and
                  Liu, Pengfei  and
                  Zhao, Yilun  and
                  Nan, Linyong  and
                  Han, Ruilin  and
                  Han, Simeng  and
                  Joty, Shafiq  and
                  Wu, Chien-Sheng  and
                  Xiong, Caiming  and
                  Radev, Dragomir},
  booktitle    = {Proceedings of the 61st Annual Meeting of the Association for Computational Linguistics (Volume 1: Long Papers)},
  doi          = {10.18653/v1/2023.acl-long.228},
  editor       = {Rogers, Anna  and
                  Boyd-Graber, Jordan  and
                  Okazaki, Naoaki},
  month        = jul,
  pages        = {4140--4170},
  publisher    = {Association for Computational Linguistics},
  title        = {Revisiting the Gold Standard: Grounding Summarization Evaluation with Robust Human Evaluation},
  url          = {https://aclanthology.org/2023.acl-long.228/},
  year         = {2023}
}

@inproceedings{chen-etal-2020-mocha,
  address      = {Online},
  author       = {Chen, Anthony  and
                  Stanovsky, Gabriel  and
                  Singh, Sameer  and
                  Gardner, Matt},
  booktitle    = {Proceedings of the 2020 Conference on Empirical Methods in Natural Language Processing (EMNLP)},
  doi          = {10.18653/v1/2020.emnlp-main.528},
  editor       = {Webber, Bonnie  and
                  Cohn, Trevor  and
                  He, Yulan  and
                  Liu, Yang},
  month        = nov,
  pages        = {6521--6532},
  publisher    = {Association for Computational Linguistics},
  title        = {{MOCHA}: A Dataset for Training and Evaluating Generative Reading Comprehension Metrics},
  url          = {https://aclanthology.org/2020.emnlp-main.528/},
  year         = {2020}
}

@inproceedings{freitag-etal-2022-results,
  address      = {Abu Dhabi, United Arab Emirates (Hybrid)},
  author       = {Freitag, Markus  and
                  Rei, Ricardo  and
                  Mathur, Nitika  and
                  Lo, Chi-kiu  and
                  Stewart, Craig  and
                  Avramidis, Eleftherios  and
                  Kocmi, Tom  and
                  Foster, George  and
                  Lavie, Alon  and
                  Martins, Andr{\'e} F. T.},
  booktitle    = {Proceedings of the Seventh Conference on Machine Translation (WMT)},
  editor       = {Koehn, Philipp  and
                  Barrault, Lo{\"i}c  and
                  Bojar, Ond{\v{r}}ej  and
                  Bougares, Fethi  and
                  Chatterjee, Rajen  and
                  Costa-juss{\`a}, Marta R.  and
                  Federmann, Christian  and
                  Fishel, Mark  and
                  Fraser, Alexander  and
                  Freitag, Markus  and
                  Graham, Yvette  and
                  Grundkiewicz, Roman  and
                  Guzman, Paco  and
                  Haddow, Barry  and
                  Huck, Matthias  and
                  Jimeno Yepes, Antonio  and
                  Kocmi, Tom  and
                  Martins, Andr{\'e}  and
                  Morishita, Makoto  and
                  Monz, Christof  and
                  Nagata, Masaaki  and
                  Nakazawa, Toshiaki  and
                  Negri, Matteo  and
                  N{\'e}v{\'e}ol, Aur{\'e}lie  and
                  Neves, Mariana  and
                  Popel, Martin  and
                  Turchi, Marco  and
                  Zampieri, Marcos},
  month        = dec,
  pages        = {46--68},
  publisher    = {Association for Computational Linguistics},
  title        = {{R}esults of {WMT}22 {M}etrics {S}hared {T}ask: {S}top {U}sing {BLEU} {--} {N}eural {M}etrics {A}re {B}etter and {M}ore {R}obust},
  url          = {https://aclanthology.org/2022.wmt-1.2/},
  year         = {2022}
}

@article{libovicky-etal-2025-cus,
  author       = {Jindrich Libovick{\'{y}} and
                  Jindrich Helcl and
                  Andrei Manea and
                  Gianluca Vico},
  bibsource    = {dblp computer science bibliography, https://dblp.org},
  doi          = {10.48550/ARXIV.2507.22752},
  eprint       = {2507.22752},
  eprinttype   = {arXiv},
  journal      = {CoRR},
  title        = {{CUS-QA:} {L}ocal-{K}nowledge-Oriented {O}pen-Ended {Q}uestion {A}nswering {D}ataset},
  url          = {https://doi.org/10.48550/arXiv.2507.22752},
  volume       = {abs/2507.22752},
  year         = {2025}
}

@inproceedings{sellam-etal-2020-bleurt,
  address      = {Online},
  author       = {Sellam, Thibault  and
                  Das, Dipanjan  and
                  Parikh, Ankur},
  booktitle    = {Proceedings of the 58th Annual Meeting of the Association for Computational Linguistics},
  doi          = {10.18653/v1/2020.acl-main.704},
  editor       = {Jurafsky, Dan  and
                  Chai, Joyce  and
                  Schluter, Natalie  and
                  Tetreault, Joel},
  month        = jul,
  pages        = {7881--7892},
  publisher    = {Association for Computational Linguistics},
  title        = {{BLEURT}: Learning Robust Metrics for Text Generation},
  url          = {https://aclanthology.org/2020.acl-main.704/},
  year         = {2020}
}

@inproceedings{akhbardeh-etal-2021-findings,
  address      = {Online},
  author       = {Akhbardeh, Farhad  and
                  Arkhangorodsky, Arkady  and
                  Biesialska, Magdalena  and
                  Bojar, Ond{\v{r}}ej  and
                  Chatterjee, Rajen  and
                  Chaudhary, Vishrav  and
                  Costa-jussa, Marta R.  and
                  Espa{\~n}a-Bonet, Cristina  and
                  Fan, Angela  and
                  Federmann, Christian  and
                  Freitag, Markus  and
                  Graham, Yvette  and
                  Grundkiewicz, Roman  and
                  Haddow, Barry  and
                  Harter, Leonie  and
                  Heafield, Kenneth  and
                  Homan, Christopher  and
                  Huck, Matthias  and
                  Amponsah-Kaakyire, Kwabena  and
                  Kasai, Jungo  and
                  Khashabi, Daniel  and
                  Knight, Kevin  and
                  Kocmi, Tom  and
                  Koehn, Philipp  and
                  Lourie, Nicholas  and
                  Monz, Christof  and
                  Morishita, Makoto  and
                  Nagata, Masaaki  and
                  Nagesh, Ajay  and
                  Nakazawa, Toshiaki  and
                  Negri, Matteo  and
                  Pal, Santanu  and
                  Tapo, Allahsera Auguste  and
                  Turchi, Marco  and
                  Vydrin, Valentin  and
                  Zampieri, Marcos},
  booktitle    = {Proceedings of the Sixth Conference on Machine Translation},
  editor       = {Barrault, Loic  and
                  Bojar, Ondrej  and
                  Bougares, Fethi  and
                  Chatterjee, Rajen  and
                  Costa-jussa, Marta R.  and
                  Federmann, Christian  and
                  Fishel, Mark  and
                  Fraser, Alexander  and
                  Freitag, Markus  and
                  Graham, Yvette  and
                  Grundkiewicz, Roman  and
                  Guzman, Paco  and
                  Haddow, Barry  and
                  Huck, Matthias  and
                  Yepes, Antonio Jimeno  and
                  Koehn, Philipp  and
                  Kocmi, Tom  and
                  Martins, Andre  and
                  Morishita, Makoto  and
                  Monz, Christof},
  month        = nov,
  pages        = {1--88},
  publisher    = {Association for Computational Linguistics},
  title        = {{F}indings of the 2021 {C}onference on {M}achine {T}ranslation ({WMT}21)},
  url          = {https://aclanthology.org/2021.wmt-1.1/},
  year         = {2021}
}

@inproceedings{kocmi-etal-2024-findings,
  address      = {Miami, Florida, USA},
  author       = {Kocmi, Tom  and
                  Avramidis, Eleftherios  and
                  Bawden, Rachel  and
                  Bojar, Ond{\v{r}}ej  and
                  Dvorkovich, Anton  and
                  Federmann, Christian  and
                  Fishel, Mark  and
                  Freitag, Markus  and
                  Gowda, Thamme  and
                  Grundkiewicz, Roman  and
                  Haddow, Barry  and
                  Karpinska, Marzena  and
                  Koehn, Philipp  and
                  Marie, Benjamin  and
                  Monz, Christof  and
                  Murray, Kenton  and
                  Nagata, Masaaki  and
                  Popel, Martin  and
                  Popovi{\'c}, Maja  and
                  Shmatova, Mariya  and
                  Steingr{\'i}msson, Steinth{\'o}r  and
                  Zouhar, Vil{\'e}m},
  booktitle    = {Proceedings of the Ninth Conference on Machine Translation},
  doi          = {10.18653/v1/2024.wmt-1.1},
  editor       = {Haddow, Barry  and
                  Kocmi, Tom  and
                  Koehn, Philipp  and
                  Monz, Christof},
  month        = nov,
  pages        = {1--46},
  publisher    = {Association for Computational Linguistics},
  title        = {{F}indings of the {WMT}24 {G}eneral {M}achine {T}ranslation {S}hared {T}ask: {T}he {LLM} {E}ra {I}s {H}ere but {MT} {I}s {N}ot {S}olved {Y}et},
  url          = {https://aclanthology.org/2024.wmt-1.1/},
  year         = {2024}
}

@misc{llama4,
  author       = {{Meta AI}},
  howpublished = {\url{https://ai.meta.com/blog/llama-4-multimodal-intelligence/}},
  title        = {{T}he {L}lama 4 {H}erd: {T}he {B}eginning of a {N}ew {E}ra of {N}atively {M}ultimodal {A}{I} {I}nnovation},
  year         = {2025}
}

@article{grattafiori-etal-2024-llama3,
  author       = {Llama Team},
  bibsource    = {dblp computer science bibliography, https://dblp.org},
  doi          = {10.48550/ARXIV.2407.21783},
  eprint       = {2407.21783},
  eprinttype   = {arXiv},
  journal      = {CoRR},
  title        = {{T}he {L}lama 3 {H}erd of {M}odels},
  url          = {https://doi.org/10.48550/arXiv.2407.21783},
  volume       = {abs/2407.21783},
  year         = {2024}
}

@article{qwen3,
  author       = {An Yang and
                  Anfeng Li and
                  Baosong Yang and
                  Beichen Zhang and
                  Binyuan Hui and
                  Bo Zheng and
                  Bowen Yu and
                  Chang Gao and
                  Chengen Huang and
                  Chenxu Lv and
                  Chujie Zheng and
                  Dayiheng Liu and
                  Fan Zhou and
                  Fei Huang and
                  Feng Hu and
                  Hao Ge and
                  Haoran Wei and
                  Huan Lin and
                  Jialong Tang and
                  Jian Yang and
                  Jianhong Tu and
                  Jianwei Zhang and
                  Jian Yang and
                  Jiaxi Yang and
                  Jingren Zhou and
                  Junyang Lin and
                  Kai Dang and
                  Keqin Bao and
                  Kexin Yang and
                  Le Yu and
                  Lianghao Deng and
                  Mei Li and
                  Mingfeng Xue and
                  Mingze Li and
                  Pei Zhang and
                  Peng Wang and
                  Qin Zhu and
                  Rui Men and
                  Ruize Gao and
                  Shixuan Liu and
                  Shuang Luo and
                  Tianhao Li and
                  Tianyi Tang and
                  Wenbiao Yin and
                  Xingzhang Ren and
                  Xinyu Wang and
                  Xinyu Zhang and
                  Xuancheng Ren and
                  Yang Fan and
                  Yang Su and
                  Yichang Zhang and
                  Yinger Zhang and
                  Yu Wan and
                  Yuqiong Liu and
                  Zekun Wang and
                  Zeyu Cui and
                  Zhenru Zhang and
                  Zhipeng Zhou and
                  Zihan Qiu},
  bibsource    = {dblp computer science bibliography, https://dblp.org},
  doi          = {10.48550/ARXIV.2505.09388},
  eprint       = {2505.09388},
  eprinttype   = {arXiv},
  journal      = {CoRR},
  title        = {{Q}wen3 {T}echnical {R}eport},
  url          = {https://doi.org/10.48550/arXiv.2505.09388},
  volume       = {abs/2505.09388},
  year         = {2025}
}

\appendix

\section{Metric Parameters}\label{app:parameters}

We use four string overlap metrics: BLEU with smoothing and $n$-gram orders 1--4, chrF with character order $c \in \{4,6\}$ and word order $w \in \{0,2\}$, ROUGE-1, -2, -4, and -L, and METEOR with $\alpha \in \{0.2, 0.9\}$ and $\gamma \in \{0, 0.5\}$.

For embedding-based metrics, we compute BERTScore using Czech and English models with limited (5 layers) and full depth.

For model-based metrics, we use four COMET models (\texttt{Unbabel/wmt22-comet-da}, \texttt{eamt22-cometinho-da}, \texttt{Unbabel/wmt20-comet-da}, \texttt{Unbabel/wmt20-comet-qe-da}) and four BLEURT checkpoints (\texttt{bleurt-tiny-128}, \texttt{bleurt-base-512}, \texttt{bleurt-large-512}, \texttt{BLEURT-20-D12}).

\section{Prompts for Synthetic Data Generation}
\label{sec:prompts}

Tables~\ref{tab:prompt_cusqa_zero}--\ref{tab:prompt_mt_few} present the prompts used to generate synthetic data. We provide zero-shot and few-shot variants for each task: CUS-QA, MOCHA, RoSE, and machine translation. All prompts define six damage levels (0--5) with task-specific degradation strategies.

\section{Additional Results}
\label{sec:appendix}

Table~\ref{tab:kendall} reports meta-correlation results using Kendall rank correlation, complementing the Spearman results in Table~\ref{tab:spearman}; patterns are consistent across both measures.

Tables~\ref{tab:qa_cz_combined_spearman}--\ref{tab:wmt24_enis_spearman} present detailed Spearman correlations between metric scores and either human judgments (column Hum) or synthetic damage levels for each Meta-Judge model under few-shot (F) and zero-shot (Z) prompting. These tables allow examination of individual metric behavior across datasets and languages.

Table~\ref{tab:qa_cz_combined_spearman} covers CUS-QA Czech, Table~\ref{tab:qa_uk_combined_spearman} covers CUS-QA Slovak and Ukrainian, Table~\ref{tab:wmt21_combined_spearman} presents MOCHA, RoSE, and WMT 2021, and Table~\ref{tab:wmt24_enis_spearman} reports WMT 2024 language pairs.

\section{Damage Specification Robustness}
\label{sec:app_robustness}

Tables~\ref{tab:prompt_sensitivity_v1}--\ref{tab:prompt_sensitivity_v5} show the five damage specification variants used in the prompt robustness testing in Section~\ref{sec:robustness}. Apart from the damage specification, they are same as in Table~\ref{tab:prompt_cusqa_zero} and Table~\ref{tab:prompt_cusqa_few}

\begin{table*}[ht]
\footnotesize\centering
\setlength{\tabcolsep}{4pt}

\begin{tabular}{l l ccc ccc c c cc ccc}
\toprule
\multirow{2}{*}{Model} & \multirow{2}{*}{Shot} & \multicolumn{3}{c}{CUS-QA (en)} & \multicolumn{3}{c}{CUS-QA (orig.)} & \multirow{2}{*}{\textls[-60]{MOCHA}} & \multirow{2}{*}{RoSE} & \multicolumn{2}{c}{WMT 21} & \multicolumn{3}{c}{WMT 24} \\
\cmidrule(lr){3-5} \cmidrule(lr){6-8} \cmidrule(lr){11-12} \cmidrule(lr){13-15}
 & & cs & sk & uk & cs & sk & uk & & & en-ha & xh-zu & cs-uk & en-cs & en-is \\
\midrule
\multirow{2}{*}{Llama 4 Scout} & Few & \R{.730}{.0000} & \R{.661}{.0000} & \R{.492}{.0001} & \R{.751}{.0000} & \R{.735}{.0000} & \R{.533}{.0000} & \R{.249}{.0329} & \R{.365}{.0030} & \R{.312}{.0099} & \R{.265}{.0249} & \R{.825}{.0000} & \R{.148}{.1400} & \R{.233}{.0428} \\
 & Zero & \R{.656}{.0000} & \R{.624}{.0000} & \R{.450}{.0003} & \R{.688}{.0000} & \R{.704}{.0000} & \R{.573}{.0000} & \R{.138}{.1584} & \R{.365}{.0030} & \R{.201}{.0697} & \R{.318}{.0089} & \R{.762}{.0000} & \R{.318}{.0089} & \R{.302}{.0123} \\
\midrule
\multirow{2}{*}{Llama 3.3 70B} & Few & \R{.619}{.0000} & \R{.598}{.0000} & \R{.524}{.0000} & \R{.757}{.0000} & \R{.730}{.0000} & \R{.584}{.0000} & \R{.487}{.0001} & \R{.064}{.3262} & \R{.333}{.0063} & \R{.323}{.0079} & \R{.656}{.0000} & \R{.238}{.0393} & \R{.286}{.0168} \\
 & Zero & \R{.524}{.0000} & \R{.519}{.0000} & \R{.355}{.0039} & \R{.672}{.0000} & \R{.656}{.0000} & \R{.474}{.0003} & \R{.265}{.0249} & \R{.280}{.0186} & \R{.386}{.0018} & \R{.323}{.0079} & \R{.656}{.0000} & \R{.307}{.0110} & \R{.360}{.0034} \\
\midrule
\multirow{2}{*}{Qwen 3 30B} & Few & \R{.651}{.0000} & \R{.619}{.0000} & \R{.429}{.0006} & \R{.693}{.0000} & \R{.709}{.0000} & \R{.516}{.0000} & \R{.571}{.0000} & \R{.635}{.0000} & \R{.032}{.4148} & \R{-.064}{.6878} & \R{.767}{.0000} & \R{.153}{.1314} & \R{.249}{.0329} \\
 & Zero & \R{.725}{.0000} & \R{.513}{.0000} & \R{.577}{.0000} & \R{.804}{.0000} & \R{.788}{.0000} & \R{.681}{.0000} & \R{.720}{.0000} & \R{.513}{.0000} & \R{-.021}{.5699} & \R{.111}{.2107} & \R{.810}{.0000} & \R{.085}{.2718} & \R{.201}{.0697} \\
\bottomrule
\end{tabular}

\caption{Meta-correlation between LLM as Meta-Judge and standard metric validation with human judgment measured by \emph{Kendall correlation}. The values are black if they are significant at the 0.01 confidence, \textcolor{black!60}{dark gray} if they are only significant at 0.05 confidence, and \textcolor{black!20}{light gray} if they are not significantly different from zero.}
\label{tab:kendall}
\end{table*}

\begin{table*}[ht]
\footnotesize\centering
\setlength{\tabcolsep}{3.5pt}
\renewcommand{\arraystretch}{0.85}
\begin{tabular}{ll c cc cc cc @{\hspace{3mm}} c cc cc cc}
\toprule
& & \multicolumn{7}{c}{\textbf{CUS-QA cs (en)}} & \multicolumn{7}{c}{\textbf{CUS-QA cs (orig.)}} \\
\cmidrule(lr){3-9} \cmidrule(lr){10-16}
\multirow{2}{*}{\rotatebox{90}{~~~~Metric}} & \multirow{2}{*}{Parameters} & \multirow{2}{*}{Hum} & \multicolumn{2}{c}{Llama 4} & \multicolumn{2}{c}{Llama 3.3} & \multicolumn{2}{c}{Qwen 3} & \multirow{2}{*}{GT} & \multicolumn{2}{c}{Llama 4} & \multicolumn{2}{c}{Llama 3.3} & \multicolumn{2}{c}{Qwen 3} \\
\cmidrule(lr){4-5} \cmidrule(lr){6-7} \cmidrule(lr){8-9} \cmidrule(lr){11-12} \cmidrule(lr){13-14} \cmidrule(lr){15-16}
 & & & F & Z & F & Z & F & Z & & F & Z & F & Z & F & Z \\
\midrule
\multirow{4}{*}{\rotatebox{90}{BLEU}} 
 & Order 1 & \Cor{.512} & \Cor{.132} & \Cor{-.035} & \Cor{.049} & \Cor{-.154} & \Cor{.048} & \Cor{.080} & \Cor{.478} & \Cor{.091} & \Cor{-.061} & \Cor{.099} & \Cor{-.103} & \Cor{.007} & \Cor{.134} \\
 & Order 2 & \Cor{.493} & \Cor{.112} & \Cor{-.088} & \Cor{.026} & \Cor{-.229} & \Cor{.022} & \Cor{.040} & \Cor{.451} & \Cor{.073} & \Cor{-.115} & \Cor{.095} & \Cor{-.183} & \Cor{-.006} & \Cor{.109} \\
 & Order 3 & \Cor{.467} & \Cor{.064} & \Cor{-.141} & \Cor{-.012} & \Cor{-.285} & \Cor{-.018} & \Cor{.018} & \Cor{.394} & \Cor{.003} & \Cor{-.193} & \Cor{.052} & \Cor{-.255} & \Cor{-.054} & \Cor{.060} \\
 & Order 4 & \Cor{.447} & \Cor{.025} & \Cor{-.173} & \Cor{-.044} & \Cor{-.318} & \Cor{-.048} & \Cor{.009} & \Cor{.343} & \Cor{-.052} & \Cor{-.238} & \Cor{.015} & \Cor{-.295} & \Cor{-.083} & \Cor{.029} \\
\midrule
\multirow{4}{*}{\rotatebox{90}{chrF}} 
 & c: 4, w: 0 & \Cor{.617} & \Cor{.487} & \Cor{.429} & \Cor{.420} & \Cor{.352} & \Cor{.425} & \Cor{.363} & \Cor{.655} & \Cor{.505} & \Cor{.433} & \Cor{.452} & \Cor{.418} & \Cor{.436} & \Cor{.488} \\
 & c: 4, w: 2 & \Cor{.626} & \Cor{.495} & \Cor{.427} & \Cor{.409} & \Cor{.338} & \Cor{.422} & \Cor{.351} & \Cor{.667} & \Cor{.515} & \Cor{.441} & \Cor{.453} & \Cor{.417} & \Cor{.446} & \Cor{.498} \\
 & c: 6, w: 0 & \Cor{.624} & \Cor{.510} & \Cor{.452} & \Cor{.444} & \Cor{.372} & \Cor{.451} & \Cor{.374} & \Cor{.665} & \Cor{.522} & \Cor{.452} & \Cor{.467} & \Cor{.437} & \Cor{.458} & \Cor{.502} \\
 & c: 6, w: 2 & \Cor{.630} & \Cor{.512} & \Cor{.445} & \Cor{.430} & \Cor{.358} & \Cor{.445} & \Cor{.364} & \Cor{.672} & \Cor{.527} & \Cor{.455} & \Cor{.466} & \Cor{.432} & \Cor{.462} & \Cor{.507} \\
\midrule
\multirow{4}{*}{\rotatebox{90}{ROUGE}} 
 & ROUGE-1 & \Cor{.615} & \Cor{.390} & \Cor{.338} & \Cor{.328} & \Cor{.241} & \Cor{.321} & \Cor{.258} & \Cor{.626} & \Cor{.385} & \Cor{.295} & \Cor{.334} & \Cor{.275} & \Cor{.280} & \Cor{.361} \\
 & ROUGE-2 & \Cor{.476} & \Cor{.405} & \Cor{.362} & \Cor{.336} & \Cor{.283} & \Cor{.352} & \Cor{.232} & \Cor{.580} & \Cor{.415} & \Cor{.348} & \Cor{.339} & \Cor{.318} & \Cor{.324} & \Cor{.360} \\
 & ROUGE-4 & \Cor{.259} & \Cor{.230} & \Cor{.204} & \Cor{.173} & \Cor{.120} & \Cor{.182} & \Cor{.070} & \Cor{.377} & \Cor{.267} & \Cor{.241} & \Cor{.215} & \Cor{.191} & \Cor{.211} & \Cor{.227} \\
 & ROUGE-L & \Cor{.619} & \Cor{.385} & \Cor{.334} & \Cor{.319} & \Cor{.233} & \Cor{.319} & \Cor{.256} & \Cor{.629} & \Cor{.387} & \Cor{.294} & \Cor{.330} & \Cor{.270} & \Cor{.281} & \Cor{.361} \\
\midrule
\multirow{4}{*}{\rotatebox{90}{\textls[-40]{METEOR}}} 
 & $\alpha$: 0.2, $\gamma$: 0.0 & \Cor{.552} & \Cor{.247} & \Cor{.179} & \Cor{.186} & \Cor{.129} & \Cor{.196} & \Cor{.233} & \Cor{.553} & \Cor{.227} & \Cor{.146} & \Cor{.187} & \Cor{.139} & \Cor{.150} & \Cor{.270} \\
 & $\alpha$: 0.2, $\gamma$: 0.5 & \Cor{.564} & \Cor{.303} & \Cor{.232} & \Cor{.237} & \Cor{.168} & \Cor{.246} & \Cor{.225} & \Cor{.576} & \Cor{.309} & \Cor{.225} & \Cor{.255} & \Cor{.200} & \Cor{.227} & \Cor{.314} \\
 & $\alpha$: 0.9, $\gamma$: 0.0 & \Cor{.593} & \Cor{.462} & \Cor{.408} & \Cor{.394} & \Cor{.329} & \Cor{.429} & \Cor{.274} & \Cor{.621} & \Cor{.491} & \Cor{.423} & \Cor{.411} & \Cor{.383} & \Cor{.442} & \Cor{.464} \\
 & $\alpha$: 0.9, $\gamma$: 0.5 & \Cor{.580} & \Cor{.459} & \Cor{.398} & \Cor{.393} & \Cor{.325} & \Cor{.423} & \Cor{.249} & \Cor{.616} & \Cor{.492} & \Cor{.423} & \Cor{.418} & \Cor{.391} & \Cor{.443} & \Cor{.453} \\
\midrule
\multirow{4}{*}{\rotatebox{90}{BERTSc.}} 
 & cs, Not Limited & \Cor{.590} & \Cor{.342} & \Cor{.243} & \Cor{.271} & \Cor{.127} & \Cor{.270} & \Cor{.273} & \Cor{.609} & \Cor{.354} & \Cor{.236} & \Cor{.315} & \Cor{.180} & \Cor{.257} & \Cor{.342} \\
 & cs, Limited (L5) & \Cor{.509} & \Cor{.172} & \Cor{.030} & \Cor{.130} & \Cor{-.070} & \Cor{.110} & \Cor{.156} & \Cor{.520} & \Cor{.166} & \Cor{.020} & \Cor{.174} & \Cor{-.028} & \Cor{.081} & \Cor{.191} \\
 & en, Not Limited & \Cor{.598} & \Cor{.481} & \Cor{.422} & \Cor{.439} & \Cor{.368} & \Cor{.425} & \Cor{.311} & \Cor{.613} & \Cor{.398} & \Cor{.324} & \Cor{.360} & \Cor{.289} & \Cor{.319} & \Cor{.384} \\
 & en, Limited (L5) & \Cor{.547} & \Cor{.338} & \Cor{.298} & \Cor{.275} & \Cor{.215} & \Cor{.294} & \Cor{.226} & \Cor{.532} & \Cor{.264} & \Cor{.207} & \Cor{.243} & \Cor{.144} & \Cor{.195} & \Cor{.281} \\
\midrule
\multirow{4}{*}{\rotatebox{90}{COMET}} 
 & wmt20-da & \Cor{.481} & \Cor{.077} & \Cor{-.005} & \Cor{.026} & \Cor{-.117} & \Cor{.045} & \Cor{.198} & \Cor{.534} & \Cor{.150} & \Cor{.059} & \Cor{.167} & \Cor{.047} & \Cor{.108} & \Cor{.230} \\
 & wmt20-qe-da & \Cor{.013} & \Cor{.089} & \Cor{.184} & \Cor{.182} & \Cor{.374} & \Cor{.133} & \Cor{.018} & \Cor{.102} & \Cor{.170} & \Cor{.275} & \Cor{.148} & \Cor{.372} & \Cor{.207} & \Cor{.125} \\
 & wmt22-da & \Cor{.530} & \Cor{.115} & \Cor{.035} & \Cor{.087} & \Cor{-.063} & \Cor{.092} & \Cor{.221} & \Cor{.537} & \Cor{.195} & \Cor{.129} & \Cor{.203} & \Cor{.129} & \Cor{.153} & \Cor{.271} \\
 & cometinho-da & \Cor{.539} & \Cor{.138} & \Cor{.053} & \Cor{.087} & \Cor{-.047} & \Cor{.107} & \Cor{.189} & \Cor{.526} & \Cor{.165} & \Cor{.080} & \Cor{.180} & \Cor{.060} & \Cor{.110} & \Cor{.229} \\
\midrule
\multirow{4}{*}{\rotatebox{90}{\textls[-30]{BLEURT}}} 
 & 20-D12 & \Cor{.595} & \Cor{.365} & \Cor{.272} & \Cor{.319} & \Cor{.183} & \Cor{.287} & \Cor{.339} & \Cor{.626} & \Cor{.370} & \Cor{.269} & \Cor{.356} & \Cor{.270} & \Cor{.310} & \Cor{.387} \\
 & base-512 & \Cor{.474} & \Cor{.119} & \Cor{.027} & \Cor{.056} & \Cor{-.062} & \Cor{.053} & \Cor{.081} & \Cor{.358} & \Cor{-.044} & \Cor{-.102} & \Cor{-.045} & \Cor{-.150} & \Cor{-.072} & \Cor{.098} \\
 & large-512 & \Cor{.536} & \Cor{.292} & \Cor{.180} & \Cor{.277} & \Cor{.152} & \Cor{.214} & \Cor{.114} & \Cor{.442} & \Cor{.057} & \Cor{-.049} & \Cor{.084} & \Cor{-.061} & \Cor{.015} & \Cor{.131} \\
 & tiny-128 & \Cor{.608} & \Cor{.436} & \Cor{.317} & \Cor{.376} & \Cor{.205} & \Cor{.350} & \Cor{.251} & \Cor{.615} & \Cor{.407} & \Cor{.260} & \Cor{.386} & \Cor{.187} & \Cor{.328} & \Cor{.373} \\
\bottomrule
\end{tabular}
\caption{Spearman Correlations across different metrics and model variations for CUS-QA cs (en) and CUS-QA cs (orig.).}
\label{tab:qa_cz_combined_spearman}
\end{table*}

\begin{table*}[ht]
\footnotesize\centering
\setlength{\tabcolsep}{3pt}
\renewcommand{\arraystretch}{0.85}
\begin{tabular}{ll c cc cc cc @{\hspace{3mm}} c cc cc cc}
\toprule
& & \multicolumn{7}{c}{\textbf{CUS-QA sk (en)}} & \multicolumn{7}{c}{\textbf{CUS-QA sk (orig.)}} \\
\cmidrule(lr){3-9} \cmidrule(lr){10-16}
\multirow{2}{*}{\rotatebox{90}{~~~~Metric}} & \multirow{2}{*}{Parameters} & \multirow{2}{*}{Hum} & \multicolumn{2}{c}{Llama 4} & \multicolumn{2}{c}{Llama 3.3} & \multicolumn{2}{c}{Qwen 3} & \multirow{2}{*}{GT} & \multicolumn{2}{c}{Llama 4} & \multicolumn{2}{c}{Llama 3.3} & \multicolumn{2}{c}{Qwen 3} \\
\cmidrule(lr){4-5} \cmidrule(lr){6-7} \cmidrule(lr){8-9} \cmidrule(lr){11-12} \cmidrule(lr){13-14} \cmidrule(lr){15-16}
 & & & F & Z & F & Z & F & Z & & F & Z & F & Z & F & Z \\
\midrule
\multirow{4}{*}{\rotatebox{90}{BLEU}} 
 & Order 1 & \Cor{.459} & \Cor{.123} & \Cor{-.020} & \Cor{.028} & \Cor{-.134} & \Cor{.019} & \Cor{.037} & \Cor{.407} & \Cor{.071} & \Cor{-.026} & \Cor{.052} & \Cor{-.151} & \Cor{.004} & \Cor{.126} \\
 & Order 2 & \Cor{.430} & \Cor{.098} & \Cor{-.083} & \Cor{.010} & \Cor{-.208} & \Cor{.004} & \Cor{.026} & \Cor{.378} & \Cor{.053} & \Cor{-.079} & \Cor{.046} & \Cor{-.219} & \Cor{-.017} & \Cor{.093} \\
 & Order 3 & \Cor{.408} & \Cor{.037} & \Cor{-.150} & \Cor{-.036} & \Cor{-.269} & \Cor{-.045} & \Cor{.004} & \Cor{.331} & \Cor{-.021} & \Cor{-.158} & \Cor{-.004} & \Cor{-.287} & \Cor{-.074} & \Cor{.037} \\
 & Order 4 & \Cor{.390} & \Cor{-.009} & \Cor{-.193} & \Cor{-.076} & \Cor{-.300} & \Cor{-.077} & \Cor{-.010} & \Cor{.290} & \Cor{-.077} & \Cor{-.207} & \Cor{-.044} & \Cor{-.325} & \Cor{-.110} & \Cor{.005} \\
\midrule
\multirow{4}{*}{\rotatebox{90}{chrF}} 
 & c: 4, w: 0 & \Cor{.565} & \Cor{.476} & \Cor{.419} & \Cor{.417} & \Cor{.355} & \Cor{.446} & \Cor{.395} & \Cor{.575} & \Cor{.494} & \Cor{.453} & \Cor{.415} & \Cor{.380} & \Cor{.474} & \Cor{.487} \\
 & c: 4, w: 2 & \Cor{.583} & \Cor{.494} & \Cor{.430} & \Cor{.422} & \Cor{.352} & \Cor{.449} & \Cor{.368} & \Cor{.602} & \Cor{.517} & \Cor{.472} & \Cor{.425} & \Cor{.377} & \Cor{.491} & \Cor{.505} \\
 & c: 6, w: 0 & \Cor{.577} & \Cor{.494} & \Cor{.438} & \Cor{.437} & \Cor{.374} & \Cor{.468} & \Cor{.398} & \Cor{.591} & \Cor{.510} & \Cor{.470} & \Cor{.426} & \Cor{.391} & \Cor{.491} & \Cor{.500} \\
 & c: 6, w: 2 & \Cor{.588} & \Cor{.506} & \Cor{.443} & \Cor{.438} & \Cor{.368} & \Cor{.468} & \Cor{.379} & \Cor{.608} & \Cor{.525} & \Cor{.483} & \Cor{.434} & \Cor{.387} & \Cor{.503} & \Cor{.514} \\
\midrule
\multirow{4}{*}{\rotatebox{90}{ROUGE}} 
 & ROUGE-1 & \Cor{.596} & \Cor{.402} & \Cor{.357} & \Cor{.301} & \Cor{.220} & \Cor{.312} & \Cor{.205} & \Cor{.540} & \Cor{.372} & \Cor{.330} & \Cor{.288} & \Cor{.207} & \Cor{.315} & \Cor{.360} \\
 & ROUGE-2 & \Cor{.454} & \Cor{.414} & \Cor{.364} & \Cor{.345} & \Cor{.284} & \Cor{.356} & \Cor{.216} & \Cor{.502} & \Cor{.410} & \Cor{.379} & \Cor{.313} & \Cor{.268} & \Cor{.350} & \Cor{.353} \\
 & ROUGE-4 & \Cor{.184} & \Cor{.178} & \Cor{.159} & \Cor{.135} & \Cor{.112} & \Cor{.152} & \Cor{.061} & \Cor{.253} & \Cor{.213} & \Cor{.203} & \Cor{.150} & \Cor{.124} & \Cor{.191} & \Cor{.197} \\
 & ROUGE-L & \Cor{.597} & \Cor{.399} & \Cor{.353} & \Cor{.294} & \Cor{.211} & \Cor{.309} & \Cor{.205} & \Cor{.543} & \Cor{.372} & \Cor{.330} & \Cor{.286} & \Cor{.200} & \Cor{.315} & \Cor{.357} \\
\midrule
\multirow{4}{*}{\rotatebox{90}{\textls[-40]{METEOR}}} 
 & $\alpha$: 0.2, $\gamma$: 0.0 & \Cor{.536} & \Cor{.244} & \Cor{.173} & \Cor{.161} & \Cor{.108} & \Cor{.159} & \Cor{.140} & \Cor{.468} & \Cor{.210} & \Cor{.157} & \Cor{.153} & \Cor{.053} & \Cor{.143} & \Cor{.253} \\
 & $\alpha$: 0.2, $\gamma$: 0.5 & \Cor{.546} & \Cor{.320} & \Cor{.248} & \Cor{.240} & \Cor{.162} & \Cor{.241} & \Cor{.152} & \Cor{.496} & \Cor{.319} & \Cor{.265} & \Cor{.234} & \Cor{.137} & \Cor{.256} & \Cor{.309} \\
 & $\alpha$: 0.9, $\gamma$: 0.0 & \Cor{.585} & \Cor{.511} & \Cor{.466} & \Cor{.433} & \Cor{.367} & \Cor{.488} & \Cor{.239} & \Cor{.547} & \Cor{.508} & \Cor{.469} & \Cor{.389} & \Cor{.333} & \Cor{.517} & \Cor{.501} \\
 & $\alpha$: 0.9, $\gamma$: 0.5 & \Cor{.572} & \Cor{.500} & \Cor{.448} & \Cor{.431} & \Cor{.356} & \Cor{.468} & \Cor{.217} & \Cor{.541} & \Cor{.502} & \Cor{.462} & \Cor{.392} & \Cor{.335} & \Cor{.500} & \Cor{.466} \\
\midrule
\multirow{4}{*}{\rotatebox{90}{BERTSc.}} 
 & cs, Not Limited & \Cor{.548} & \Cor{.322} & \Cor{.219} & \Cor{.250} & \Cor{.128} & \Cor{.249} & \Cor{.232} & \Cor{.523} & \Cor{.324} & \Cor{.262} & \Cor{.273} & \Cor{.126} & \Cor{.284} & \Cor{.332} \\
 & cs, Limited (L5) & \Cor{.460} & \Cor{.147} & \Cor{.020} & \Cor{.103} & \Cor{-.056} & \Cor{.078} & \Cor{.125} & \Cor{.422} & \Cor{.140} & \Cor{.051} & \Cor{.133} & \Cor{-.080} & \Cor{.086} & \Cor{.177} \\
 & en, Not Limited & \Cor{.582} & \Cor{.479} & \Cor{.435} & \Cor{.450} & \Cor{.398} & \Cor{.470} & \Cor{.349} & \Cor{.561} & \Cor{.410} & \Cor{.368} & \Cor{.355} & \Cor{.274} & \Cor{.382} & \Cor{.426} \\
 & en, Limited (L5) & \Cor{.541} & \Cor{.336} & \Cor{.305} & \Cor{.291} & \Cor{.238} & \Cor{.328} & \Cor{.235} & \Cor{.461} & \Cor{.241} & \Cor{.199} & \Cor{.191} & \Cor{.077} & \Cor{.227} & \Cor{.298} \\
\midrule
\multirow{4}{*}{\rotatebox{90}{COMET}} 
 & wmt20-da & \Cor{.438} & \Cor{.060} & \Cor{-.038} & \Cor{.027} & \Cor{-.104} & \Cor{.051} & \Cor{.194} & \Cor{.432} & \Cor{.132} & \Cor{.092} & \Cor{.159} & \Cor{.018} & \Cor{.124} & \Cor{.208} \\
 & wmt20-qe-da & \Cor{.040} & \Cor{.087} & \Cor{.197} & \Cor{.198} & \Cor{.331} & \Cor{.138} & \Cor{.047} & \Cor{.100} & \Cor{.172} & \Cor{.238} & \Cor{.182} & \Cor{.365} & \Cor{.213} & \Cor{.137} \\
 & wmt22-da & \Cor{.486} & \Cor{.096} & \Cor{-.002} & \Cor{.077} & \Cor{-.066} & \Cor{.080} & \Cor{.219} & \Cor{.441} & \Cor{.174} & \Cor{.135} & \Cor{.180} & \Cor{.078} & \Cor{.169} & \Cor{.255} \\
 & cometinho-da & \Cor{.521} & \Cor{.116} & \Cor{.033} & \Cor{.083} & \Cor{-.045} & \Cor{.101} & \Cor{.190} & \Cor{.444} & \Cor{.153} & \Cor{.116} & \Cor{.159} & \Cor{.038} & \Cor{.124} & \Cor{.219} \\
\midrule
\multirow{4}{*}{\rotatebox{90}{\textls[-30]{BLEURT}}} 
 & 20-D12 & \Cor{.569} & \Cor{.335} & \Cor{.229} & \Cor{.305} & \Cor{.193} & \Cor{.271} & \Cor{.329} & \Cor{.538} & \Cor{.294} & \Cor{.240} & \Cor{.303} & \Cor{.174} & \Cor{.286} & \Cor{.331} \\
 & base-512 & \Cor{.379} & \Cor{.078} & \Cor{-.036} & \Cor{.036} & \Cor{-.053} & \Cor{.029} & \Cor{.034} & \Cor{.250} & \Cor{-.060} & \Cor{-.075} & \Cor{-.082} & \Cor{-.173} & \Cor{-.087} & \Cor{.081} \\
 & large-512 & \Cor{.477} & \Cor{.193} & \Cor{.094} & \Cor{.218} & \Cor{.108} & \Cor{.173} & \Cor{.074} & \Cor{.342} & \Cor{.014} & \Cor{-.025} & \Cor{.045} & \Cor{-.085} & \Cor{.006} & \Cor{.122} \\
 & tiny-128 & \Cor{.601} & \Cor{.460} & \Cor{.352} & \Cor{.384} & \Cor{.232} & \Cor{.386} & \Cor{.237} & \Cor{.559} & \Cor{.404} & \Cor{.329} & \Cor{.346} & \Cor{.160} & \Cor{.347} & \Cor{.372} \\
\bottomrule
\end{tabular}

\vspace{5pt}

%
%
\begin{tabular}{ll c cc cc cc @{\hspace{3mm}} c cc cc cc}
\toprule
& & \multicolumn{7}{c}{\textbf{CUS-QA uk (en)}} & \multicolumn{7}{c}{\textbf{CUS-QA uk (orig.)}} \\
\cmidrule(lr){3-9} \cmidrule(lr){10-16}
\multirow{2}{*}{\rotatebox{90}{~~~~Metric}} & \multirow{2}{*}{Parameters} & \multirow{2}{*}{Hum} & \multicolumn{2}{c}{Llama 4} & \multicolumn{2}{c}{Llama 3.3} & \multicolumn{2}{c}{Qwen 3} & \multirow{2}{*}{GT} & \multicolumn{2}{c}{Llama 4} & \multicolumn{2}{c}{Llama 3.3} & \multicolumn{2}{c}{Qwen 3} \\
\cmidrule(lr){4-5} \cmidrule(lr){6-7} \cmidrule(lr){8-9} \cmidrule(lr){11-12} \cmidrule(lr){13-14} \cmidrule(lr){15-16}
 & & & F & Z & F & Z & F & Z & & F & Z & F & Z & F & Z \\
\midrule
\multirow{4}{*}{\rotatebox{90}{BLEU}} 
 & Order 1 & \Cor{.426} & \Cor{.093} & \Cor{-.034} & \Cor{.115} & \Cor{-.145} & \Cor{.008} & \Cor{.041} & \Cor{.412} & \Cor{.006} & \Cor{-.019} & \Cor{.011} & \Cor{-.134} & \Cor{-.035} & \Cor{.085} \\
 & Order 2 & \Cor{.433} & \Cor{.066} & \Cor{-.086} & \Cor{.110} & \Cor{-.199} & \Cor{-.009} & \Cor{.013} & \Cor{.393} & \Cor{-.044} & \Cor{-.083} & \Cor{-.025} & \Cor{-.203} & \Cor{-.072} & \Cor{.041} \\
 & Order 3 & \Cor{.432} & \Cor{.024} & \Cor{-.133} & \Cor{.084} & \Cor{-.238} & \Cor{-.045} & \Cor{-.005} & \Cor{.365} & \Cor{-.105} & \Cor{-.137} & \Cor{-.064} & \Cor{-.260} & \Cor{-.124} & \Cor{-.011} \\
 & Order 4 & \Cor{.431} & \Cor{-.006} & \Cor{-.158} & \Cor{.059} & \Cor{-.260} & \Cor{-.074} & \Cor{-.014} & \Cor{.342} & \Cor{-.145} & \Cor{-.169} & \Cor{-.095} & \Cor{-.297} & \Cor{-.162} & \Cor{-.048} \\
\midrule
\multirow{4}{*}{\rotatebox{90}{chrF}} 
 & c: 4, w: 0 & \Cor{.514} & \Cor{.425} & \Cor{.372} & \Cor{.381} & \Cor{.295} & \Cor{.374} & \Cor{.302} & \Cor{.591} & \Cor{.429} & \Cor{.392} & \Cor{.364} & \Cor{.331} & \Cor{.385} & \Cor{.426} \\
 & c: 4, w: 2 & \Cor{.519} & \Cor{.427} & \Cor{.373} & \Cor{.376} & \Cor{.281} & \Cor{.374} & \Cor{.287} & \Cor{.601} & \Cor{.444} & \Cor{.406} & \Cor{.362} & \Cor{.333} & \Cor{.393} & \Cor{.440} \\
 & c: 6, w: 0 & \Cor{.527} & \Cor{.448} & \Cor{.392} & \Cor{.400} & \Cor{.313} & \Cor{.402} & \Cor{.312} & \Cor{.601} & \Cor{.449} & \Cor{.407} & \Cor{.376} & \Cor{.345} & \Cor{.403} & \Cor{.443} \\
 & c: 6, w: 2 & \Cor{.526} & \Cor{.445} & \Cor{.388} & \Cor{.393} & \Cor{.299} & \Cor{.396} & \Cor{.299} & \Cor{.606} & \Cor{.456} & \Cor{.415} & \Cor{.372} & \Cor{.345} & \Cor{.406} & \Cor{.450} \\
\midrule
\multirow{4}{*}{\rotatebox{90}{ROUGE}} 
 & ROUGE-1 & \Cor{.460} & \Cor{.332} & \Cor{.283} & \Cor{.278} & \Cor{.168} & \Cor{.245} & \Cor{.198} & \Cor{.138} & \Cor{.169} & \Cor{.142} & \Cor{.114} & \Cor{.104} & \Cor{.109} & \Cor{.106} \\
 & ROUGE-2 & \Cor{.221} & \Cor{.276} & \Cor{.251} & \Cor{.208} & \Cor{.152} & \Cor{.246} & \Cor{.119} & \Cor{.128} & \Cor{.096} & \Cor{.081} & \Cor{.066} & \Cor{.061} & \Cor{.066} & \Cor{.060} \\
 & ROUGE-4* & \Cor{.057} & \Cor{.100} & \Cor{.086} & \Cor{.039} & \Cor{.003} & \Cor{.084} & \Cor{.030} & nan & nan & nan & nan & nan & nan & nan \\
 & ROUGE-L & \Cor{.477} & \Cor{.323} & \Cor{.272} & \Cor{.271} & \Cor{.159} & \Cor{.241} & \Cor{.199} & \Cor{.138} & \Cor{.169} & \Cor{.142} & \Cor{.114} & \Cor{.104} & \Cor{.109} & \Cor{.106} \\
\midrule
\multirow{4}{*}{\rotatebox{90}{\textls[-40]{METEOR}}} 
 & $\alpha$: 0.2, $\gamma$: 0.0 & \Cor{.399} & \Cor{.228} & \Cor{.185} & \Cor{.202} & \Cor{.101} & \Cor{.154} & \Cor{.182} & \Cor{.465} & \Cor{.197} & \Cor{.191} & \Cor{.149} & \Cor{.091} & \Cor{.135} & \Cor{.232} \\
 & $\alpha$: 0.2, $\gamma$: 0.5 & \Cor{.398} & \Cor{.264} & \Cor{.214} & \Cor{.231} & \Cor{.134} & \Cor{.200} & \Cor{.159} & \Cor{.469} & \Cor{.252} & \Cor{.237} & \Cor{.185} & \Cor{.135} & \Cor{.194} & \Cor{.263} \\
 & $\alpha$: 0.9, $\gamma$: 0.0 & \Cor{.465} & \Cor{.441} & \Cor{.389} & \Cor{.379} & \Cor{.285} & \Cor{.408} & \Cor{.253} & \Cor{.518} & \Cor{.415} & \Cor{.377} & \Cor{.317} & \Cor{.292} & \Cor{.396} & \Cor{.417} \\
 & $\alpha$: 0.9, $\gamma$: 0.5 & \Cor{.435} & \Cor{.411} & \Cor{.356} & \Cor{.357} & \Cor{.273} & \Cor{.391} & \Cor{.204} & \Cor{.500} & \Cor{.407} & \Cor{.367} & \Cor{.304} & \Cor{.281} & \Cor{.389} & \Cor{.391} \\
\midrule
\multirow{4}{*}{\rotatebox{90}{BERTSc.}} 
 & cs, Not Limited & \Cor{.521} & \Cor{.330} & \Cor{.256} & \Cor{.317} & \Cor{.164} & \Cor{.267} & \Cor{.250} & \Cor{.578} & \Cor{.305} & \Cor{.282} & \Cor{.270} & \Cor{.181} & \Cor{.232} & \Cor{.300} \\
 & cs, Limited (L5) & \Cor{.453} & \Cor{.187} & \Cor{.086} & \Cor{.203} & \Cor{.002} & \Cor{.121} & \Cor{.164} & \Cor{.486} & \Cor{.085} & \Cor{.073} & \Cor{.109} & \Cor{-.026} & \Cor{.038} & \Cor{.143} \\
 & en, Not Limited & \Cor{.499} & \Cor{.428} & \Cor{.371} & \Cor{.413} & \Cor{.319} & \Cor{.385} & \Cor{.252} & \Cor{.431} & \Cor{.201} & \Cor{.215} & \Cor{.170} & \Cor{.135} & \Cor{.154} & \Cor{.231} \\
 & en, Limited (L5) & \Cor{.416} & \Cor{.279} & \Cor{.241} & \Cor{.256} & \Cor{.167} & \Cor{.243} & \Cor{.159} & \Cor{.456} & \Cor{.179} & \Cor{.188} & \Cor{.159} & \Cor{.103} & \Cor{.127} & \Cor{.207} \\
\midrule
\multirow{4}{*}{\rotatebox{90}{COMET}} 
 & wmt20-da & \Cor{.348} & \Cor{.091} & \Cor{.035} & \Cor{.121} & \Cor{-.048} & \Cor{.070} & \Cor{.179} & \Cor{.490} & \Cor{.129} & \Cor{.127} & \Cor{.136} & \Cor{.039} & \Cor{.104} & \Cor{.171} \\
 & wmt20-qe-da & \Cor{.083} & \Cor{.086} & \Cor{.168} & \Cor{.085} & \Cor{.280} & \Cor{.140} & \Cor{.039} & \Cor{-.024} & \Cor{.223} & \Cor{.210} & \Cor{.189} & \Cor{.327} & \Cor{.247} & \Cor{.185} \\
 & wmt22-da & \Cor{.386} & \Cor{.138} & \Cor{.079} & \Cor{.178} & \Cor{-.007} & \Cor{.104} & \Cor{.197} & \Cor{.481} & \Cor{.181} & \Cor{.178} & \Cor{.184} & \Cor{.111} & \Cor{.153} & \Cor{.223} \\
 & cometinho-da & \Cor{.420} & \Cor{.167} & \Cor{.115} & \Cor{.177} & \Cor{.040} & \Cor{.145} & \Cor{.164} & \Cor{.507} & \Cor{.114} & \Cor{.110} & \Cor{.131} & \Cor{.030} & \Cor{.086} & \Cor{.157} \\
\midrule
\multirow{4}{*}{\rotatebox{90}{\textls[-30]{BLEURT}}} 
 & 20-D12 & \Cor{.507} & \Cor{.353} & \Cor{.295} & \Cor{.376} & \Cor{.243} & \Cor{.318} & \Cor{.314} & \Cor{.584} & \Cor{.308} & \Cor{.290} & \Cor{.315} & \Cor{.226} & \Cor{.261} & \Cor{.307} \\
 & base-512 & \Cor{.332} & \Cor{.062} & \Cor{.022} & \Cor{.122} & \Cor{-.020} & \Cor{.036} & \Cor{.058} & \Cor{.217} & \Cor{-.145} & \Cor{-.114} & \Cor{-.142} & \Cor{-.218} & \Cor{-.164} & \Cor{-.033} \\
 & large-512 & \Cor{.394} & \Cor{.193} & \Cor{.147} & \Cor{.284} & \Cor{.139} & \Cor{.190} & \Cor{.093} & \Cor{.291} & \Cor{-.121} & \Cor{-.111} & \Cor{-.102} & \Cor{-.190} & \Cor{-.140} & \Cor{-.027} \\
 & tiny-128 & \Cor{.507} & \Cor{.384} & \Cor{.294} & \Cor{.377} & \Cor{.220} & \Cor{.337} & \Cor{.212} & \Cor{.347} & \Cor{.053} & \Cor{.035} & \Cor{.051} & \Cor{-.047} & \Cor{.047} & \Cor{.085} \\
\bottomrule
\end{tabular}
\caption{Spearman Correlations across different metrics and model variations for CUS-QA sk (en) and CUS-QA sk (orig.) above and CUS-QA uk (en) and CUS-QA uk (orig.) below. The row marked by * contains \texttt{nan} values for the Ukrainian text because the default ROUGE implementation uses a tokenizer that skips most Cyrillic characters.}
\label{tab:qa_uk_combined_spearman}
\end{table*}

\begin{table*}[ht]
\footnotesize\centering
\setlength{\tabcolsep}{3pt}
\renewcommand{\arraystretch}{0.85}
\begin{tabular}{ll c cc cc cc @{\hspace{3mm}} c cc cc cc}
\toprule
& & \multicolumn{7}{c}{\textbf{MOCHA Validation}} & \multicolumn{7}{c}{\textbf{RoSE CNNDM Test}} \\
\cmidrule(lr){3-9} \cmidrule(lr){10-16}
\multirow{2}{*}{\rotatebox{90}{~~~~Metric}} & \multirow{2}{*}{Parameters} & \multirow{2}{*}{Hum} & \multicolumn{2}{c}{Llama 4} & \multicolumn{2}{c}{Llama 3.3} & \multicolumn{2}{c}{Qwen 3} & \multirow{2}{*}{GT} & \multicolumn{2}{c}{Llama 4} & \multicolumn{2}{c}{Llama 3.3} & \multicolumn{2}{c}{Qwen 3} \\
\cmidrule(lr){4-5} \cmidrule(lr){6-7} \cmidrule(lr){8-9} \cmidrule(lr){11-12} \cmidrule(lr){13-14} \cmidrule(lr){15-16}
 & & & F & Z & F & Z & F & Z & & F & Z & F & Z & F & Z \\
\midrule
\multirow{4}{*}{\rotatebox{90}{BLEU}} 
 & Order 1 & \Cor{.362} & \Cor{.319} & \Cor{.387} & \Cor{.255} & \Cor{.333} & \Cor{.218} & \Cor{.411} & \Cor{.472} & \Cor{.409} & \Cor{.331} & \Cor{.257} & \Cor{.396} & \Cor{.556} & \Cor{.658} \\
 & Order 2 & \Cor{.345} & \Cor{.338} & \Cor{.408} & \Cor{.266} & \Cor{.350} & \Cor{.208} & \Cor{.393} & \Cor{.534} & \Cor{.429} & \Cor{.354} & \Cor{.237} & \Cor{.413} & \Cor{.595} & \Cor{.709} \\
 & Order 3 & \Cor{.312} & \Cor{.334} & \Cor{.402} & \Cor{.254} & \Cor{.349} & \Cor{.195} & \Cor{.386} & \Cor{.529} & \Cor{.429} & \Cor{.358} & \Cor{.222} & \Cor{.410} & \Cor{.599} & \Cor{.719} \\
 & Order 4 & \Cor{.277} & \Cor{.320} & \Cor{.389} & \Cor{.238} & \Cor{.345} & \Cor{.183} & \Cor{.381} & \Cor{.514} & \Cor{.426} & \Cor{.358} & \Cor{.209} & \Cor{.404} & \Cor{.593} & \Cor{.718} \\
\midrule
\multirow{4}{*}{\rotatebox{90}{chrF}} 
 & c: 4, w: 0 & \Cor{.527} & \Cor{.300} & \Cor{.359} & \Cor{.246} & \Cor{.286} & \Cor{.313} & \Cor{.402} & \Cor{.691} & \Cor{.511} & \Cor{.421} & \Cor{.300} & \Cor{.420} & \Cor{.619} & \Cor{.683} \\
 & c: 4, w: 2 & \Cor{.530} & \Cor{.308} & \Cor{.366} & \Cor{.259} & \Cor{.300} & \Cor{.319} & \Cor{.408} & \Cor{.705} & \Cor{.501} & \Cor{.419} & \Cor{.280} & \Cor{.426} & \Cor{.628} & \Cor{.710} \\
 & c: 6, w: 0 & \Cor{.543} & \Cor{.309} & \Cor{.367} & \Cor{.259} & \Cor{.296} & \Cor{.324} & \Cor{.404} & \Cor{.695} & \Cor{.516} & \Cor{.432} & \Cor{.289} & \Cor{.433} & \Cor{.641} & \Cor{.719} \\
 & c: 6, w: 2 & \Cor{.541} & \Cor{.312} & \Cor{.369} & \Cor{.267} & \Cor{.305} & \Cor{.326} & \Cor{.409} & \Cor{.702} & \Cor{.507} & \Cor{.428} & \Cor{.278} & \Cor{.433} & \Cor{.640} & \Cor{.727} \\
\midrule
\multirow{4}{*}{\rotatebox{90}{ROUGE}} 
 & ROUGE-1 & \Cor{.539} & \Cor{.293} & \Cor{.341} & \Cor{.235} & \Cor{.279} & \Cor{.299} & \Cor{.428} & \Cor{.600} & \Cor{.480} & \Cor{.396} & \Cor{.305} & \Cor{.434} & \Cor{.613} & \Cor{.698} \\
 & ROUGE-2 & \Cor{.375} & \Cor{.194} & \Cor{.228} & \Cor{.139} & \Cor{.156} & \Cor{.196} & \Cor{.247} & \Cor{.577} & \Cor{.449} & \Cor{.388} & \Cor{.243} & \Cor{.425} & \Cor{.628} & \Cor{.733} \\
 & ROUGE-4 & \Cor{.218} & \Cor{.058} & \Cor{.087} & \Cor{.011} & \Cor{.020} & \Cor{.062} & \Cor{.082} & \Cor{.466} & \Cor{.331} & \Cor{.303} & \Cor{.136} & \Cor{.352} & \Cor{.522} & \Cor{.659} \\
 & ROUGE-L & \Cor{.540} & \Cor{.295} & \Cor{.342} & \Cor{.238} & \Cor{.282} & \Cor{.299} & \Cor{.429} & \Cor{.558} & \Cor{.458} & \Cor{.383} & \Cor{.273} & \Cor{.435} & \Cor{.633} & \Cor{.731} \\
\midrule
\multirow{4}{*}{\rotatebox{90}{\textls[-40]{METEOR}}} 
 & $\alpha$: 0.2, $\gamma$: 0.0 & \Cor{.542} & \Cor{.286} & \Cor{.337} & \Cor{.230} & \Cor{.285} & \Cor{.267} & \Cor{.423} & \Cor{.398} & \Cor{.437} & \Cor{.349} & \Cor{.295} & \Cor{.404} & \Cor{.562} & \Cor{.652} \\
 & $\alpha$: 0.2, $\gamma$: 0.5 & \Cor{.527} & \Cor{.284} & \Cor{.334} & \Cor{.231} & \Cor{.278} & \Cor{.270} & \Cor{.398} & \Cor{.474} & \Cor{.443} & \Cor{.349} & \Cor{.264} & \Cor{.408} & \Cor{.588} & \Cor{.686} \\
 & $\alpha$: 0.9, $\gamma$: 0.0 & \Cor{.533} & \Cor{.317} & \Cor{.362} & \Cor{.270} & \Cor{.294} & \Cor{.357} & \Cor{.419} & \Cor{.712} & \Cor{.499} & \Cor{.435} & \Cor{.299} & \Cor{.451} & \Cor{.601} & \Cor{.715} \\
 & $\alpha$: 0.9, $\gamma$: 0.5 & \Cor{.516} & \Cor{.299} & \Cor{.340} & \Cor{.256} & \Cor{.275} & \Cor{.329} & \Cor{.375} & \Cor{.689} & \Cor{.483} & \Cor{.409} & \Cor{.263} & \Cor{.438} & \Cor{.617} & \Cor{.729} \\
\midrule
\multirow{4}{*}{\rotatebox{90}{BERTSc.}} 
 & cs, Not Limited & \Cor{.528} & \Cor{.315} & \Cor{.372} & \Cor{.271} & \Cor{.330} & \Cor{.276} & \Cor{.424} & \Cor{.632} & \Cor{.457} & \Cor{.396} & \Cor{.266} & \Cor{.445} & \Cor{.618} & \Cor{.733} \\
 & cs, Limited (L5) & \Cor{.440} & \Cor{.282} & \Cor{.346} & \Cor{.253} & \Cor{.318} & \Cor{.215} & \Cor{.403} & \Cor{.610} & \Cor{.406} & \Cor{.354} & \Cor{.239} & \Cor{.436} & \Cor{.589} & \Cor{.721} \\
 & en, Not Limited & \Cor{.538} & \Cor{.338} & \Cor{.382} & \Cor{.292} & \Cor{.345} & \Cor{.295} & \Cor{.403} & \Cor{.634} & \Cor{.496} & \Cor{.425} & \Cor{.292} & \Cor{.452} & \Cor{.628} & \Cor{.739} \\
 & en, Limited (L5) & \Cor{.423} & \Cor{.277} & \Cor{.333} & \Cor{.245} & \Cor{.318} & \Cor{.223} & \Cor{.392} & \Cor{.626} & \Cor{.516} & \Cor{.425} & \Cor{.393} & \Cor{.499} & \Cor{.617} & \Cor{.729} \\
\midrule
\multirow{4}{*}{\rotatebox{90}{COMET}} 
 & wmt20-da & \Cor{.620} & \Cor{.311} & \Cor{.350} & \Cor{.310} & \Cor{.356} & \Cor{.318} & \Cor{.471} & \Cor{.475} & \Cor{.438} & \Cor{.410} & \Cor{.455} & \Cor{.507} & \Cor{.520} & \Cor{.628} \\
 & wmt20-qe-da & \Cor{.015} & \Cor{-.132} & \Cor{-.129} & \Cor{-.036} & \Cor{-.048} & \Cor{-.048} & \Cor{-.229} & \Cor{.235} & \Cor{.105} & \Cor{.116} & \Cor{.217} & \Cor{.211} & \Cor{.065} & \Cor{.025} \\
 & wmt22-da & \Cor{.595} & \Cor{.318} & \Cor{.359} & \Cor{.333} & \Cor{.367} & \Cor{.313} & \Cor{.455} & \Cor{.503} & \Cor{.474} & \Cor{.444} & \Cor{.475} & \Cor{.526} & \Cor{.519} & \Cor{.643} \\
 & cometinho-da & \Cor{.570} & \Cor{.306} & \Cor{.374} & \Cor{.292} & \Cor{.370} & \Cor{.286} & \Cor{.448} & \Cor{.459} & \Cor{.403} & \Cor{.379} & \Cor{.362} & \Cor{.459} & \Cor{.529} & \Cor{.653} \\
\midrule
\multirow{4}{*}{\rotatebox{90}{\textls[-30]{BLEURT}}} 
 & 20-D12 & \Cor{.529} & \Cor{.313} & \Cor{.339} & \Cor{.311} & \Cor{.346} & \Cor{.328} & \Cor{.454} & \Cor{.633} & \Cor{.582} & \Cor{.467} & \Cor{.421} & \Cor{.431} & \Cor{.601} & \Cor{.660} \\
 & base-512 & \Cor{.582} & \Cor{.310} & \Cor{.344} & \Cor{.326} & \Cor{.372} & \Cor{.295} & \Cor{.474} & \Cor{.423} & \Cor{.536} & \Cor{.431} & \Cor{.501} & \Cor{.445} & \Cor{.491} & \Cor{.572} \\
 & large-512 & \Cor{.624} & \Cor{.393} & \Cor{.393} & \Cor{.439} & \Cor{.449} & \Cor{.412} & \Cor{.525} & \Cor{.548} & \Cor{.633} & \Cor{.525} & \Cor{.610} & \Cor{.545} & \Cor{.570} & \Cor{.636} \\
 & tiny-128 & \Cor{.576} & \Cor{.375} & \Cor{.430} & \Cor{.340} & \Cor{.424} & \Cor{.344} & \Cor{.482} & \Cor{.419} & \Cor{.582} & \Cor{.487} & \Cor{.410} & \Cor{.424} & \Cor{.540} & \Cor{.624} \\
\bottomrule
\end{tabular}

\vspace{5pt}

\begin{tabular}{ll c cc cc cc @{\hspace{3mm}} c cc cc cc}
\toprule
& & \multicolumn{7}{c}{\textbf{WMT 21 en-ha}} & \multicolumn{7}{c}{\textbf{WMT 21 xh-zu}} \\
\cmidrule(lr){3-9} \cmidrule(lr){10-16}
\multirow{2}{*}{\rotatebox{90}{~~~~Metric}} & \multirow{2}{*}{Parameters} & \multirow{2}{*}{Hum} & \multicolumn{2}{c}{Llama 4} & \multicolumn{2}{c}{Llama 3.3} & \multicolumn{2}{c}{Qwen 3} & \multirow{2}{*}{GT} & \multicolumn{2}{c}{Llama 4} & \multicolumn{2}{c}{Llama 3.3} & \multicolumn{2}{c}{Qwen 3} \\
\cmidrule(lr){4-5} \cmidrule(lr){6-7} \cmidrule(lr){8-9} \cmidrule(lr){11-12} \cmidrule(lr){13-14} \cmidrule(lr){15-16}
 & & & F & Z & F & Z & F & Z & & F & Z & F & Z & F & Z \\
\midrule
\multirow{4}{*}{\rotatebox{90}{BLEU}} 
 & Order 1 & \Cor{.174} & \Cor{.373} & \Cor{.285} & \Cor{.179} & \Cor{.200} & \Cor{.285} & \Cor{.078} & \Cor{.212} & \Cor{.349} & \Cor{.232} & \Cor{.193} & \Cor{.220} & \Cor{.349} & \Cor{.152} \\
 & Order 2 & \Cor{.175} & \Cor{.357} & \Cor{.275} & \Cor{.162} & \Cor{.185} & \Cor{.289} & \Cor{.081} & \Cor{.217} & \Cor{.331} & \Cor{.224} & \Cor{.187} & \Cor{.211} & \Cor{.355} & \Cor{.156} \\
 & Order 3 & \Cor{.166} & \Cor{.345} & \Cor{.267} & \Cor{.154} & \Cor{.177} & \Cor{.288} & \Cor{.080} & \Cor{.215} & \Cor{.314} & \Cor{.213} & \Cor{.175} & \Cor{.199} & \Cor{.355} & \Cor{.157} \\
 & Order 4 & \Cor{.160} & \Cor{.335} & \Cor{.259} & \Cor{.148} & \Cor{.170} & \Cor{.287} & \Cor{.080} & \Cor{.208} & \Cor{.298} & \Cor{.200} & \Cor{.161} & \Cor{.186} & \Cor{.355} & \Cor{.157} \\
\midrule
\multirow{4}{*}{\rotatebox{90}{chrF}} 
 & c: 4, w: 0 & \Cor{.192} & \Cor{.392} & \Cor{.292} & \Cor{.171} & \Cor{.181} & \Cor{.298} & \Cor{.091} & \Cor{.268} & \Cor{.371} & \Cor{.245} & \Cor{.200} & \Cor{.216} & \Cor{.365} & \Cor{.163} \\
 & c: 4, w: 2 & \Cor{.193} & \Cor{.376} & \Cor{.280} & \Cor{.158} & \Cor{.173} & \Cor{.289} & \Cor{.085} & \Cor{.266} & \Cor{.357} & \Cor{.238} & \Cor{.193} & \Cor{.215} & \Cor{.356} & \Cor{.156} \\
 & c: 6, w: 0 & \Cor{.196} & \Cor{.379} & \Cor{.285} & \Cor{.163} & \Cor{.174} & \Cor{.295} & \Cor{.090} & \Cor{.276} & \Cor{.363} & \Cor{.244} & \Cor{.198} & \Cor{.214} & \Cor{.361} & \Cor{.161} \\
 & c: 6, w: 2 & \Cor{.195} & \Cor{.372} & \Cor{.280} & \Cor{.157} & \Cor{.171} & \Cor{.289} & \Cor{.085} & \Cor{.273} & \Cor{.356} & \Cor{.240} & \Cor{.195} & \Cor{.214} & \Cor{.356} & \Cor{.157} \\
\midrule
\multirow{4}{*}{\rotatebox{90}{ROUGE}} 
 & ROUGE-1 & \Cor{.173} & \Cor{.371} & \Cor{.294} & \Cor{.184} & \Cor{.191} & \Cor{.293} & \Cor{.098} & \Cor{.201} & \Cor{.356} & \Cor{.259} & \Cor{.207} & \Cor{.215} & \Cor{.359} & \Cor{.156} \\
 & ROUGE-2 & \Cor{.159} & \Cor{.347} & \Cor{.275} & \Cor{.161} & \Cor{.175} & \Cor{.297} & \Cor{.097} & \Cor{.193} & \Cor{.302} & \Cor{.213} & \Cor{.151} & \Cor{.174} & \Cor{.355} & \Cor{.158} \\
 & ROUGE-4 & \Cor{.103} & \Cor{.286} & \Cor{.221} & \Cor{.105} & \Cor{.118} & \Cor{.286} & \Cor{.092} & \Cor{.135} & \Cor{.206} & \Cor{.134} & \Cor{.055} & \Cor{.091} & \Cor{.346} & \Cor{.151} \\
 & ROUGE-L & \Cor{.180} & \Cor{.388} & \Cor{.320} & \Cor{.210} & \Cor{.249} & \Cor{.305} & \Cor{.106} & \Cor{.215} & \Cor{.368} & \Cor{.277} & \Cor{.224} & \Cor{.243} & \Cor{.366} & \Cor{.172} \\
\midrule
\multirow{4}{*}{\rotatebox{90}{\textls[-40]{METEOR}}} 
 & $\alpha$: 0.2, $\gamma$: 0.0 & \Cor{.153} & \Cor{.377} & \Cor{.323} & \Cor{.228} & \Cor{.235} & \Cor{.286} & \Cor{.087} & \Cor{.195} & \Cor{.361} & \Cor{.275} & \Cor{.233} & \Cor{.237} & \Cor{.363} & \Cor{.167} \\
 & $\alpha$: 0.2, $\gamma$: 0.5 & \Cor{.158} & \Cor{.372} & \Cor{.316} & \Cor{.220} & \Cor{.231} & \Cor{.281} & \Cor{.089} & \Cor{.206} & \Cor{.359} & \Cor{.274} & \Cor{.231} & \Cor{.240} & \Cor{.359} & \Cor{.165} \\
 & $\alpha$: 0.9, $\gamma$: 0.0 & \Cor{.168} & \Cor{.364} & \Cor{.264} & \Cor{.148} & \Cor{.156} & \Cor{.269} & \Cor{.074} & \Cor{.187} & \Cor{.340} & \Cor{.221} & \Cor{.178} & \Cor{.200} & \Cor{.340} & \Cor{.140} \\
 & $\alpha$: 0.9, $\gamma$: 0.5 & \Cor{.169} & \Cor{.355} & \Cor{.261} & \Cor{.148} & \Cor{.163} & \Cor{.267} & \Cor{.078} & \Cor{.201} & \Cor{.337} & \Cor{.223} & \Cor{.182} & \Cor{.207} & \Cor{.337} & \Cor{.140} \\
\midrule
\multirow{4}{*}{\rotatebox{90}{BERTSc.}} 
 & cs, Not Limited & \Cor{.191} & \Cor{.384} & \Cor{.299} & \Cor{.193} & \Cor{.220} & \Cor{.273} & \Cor{.076} & \Cor{.257} & \Cor{.370} & \Cor{.250} & \Cor{.212} & \Cor{.224} & \Cor{.355} & \Cor{.160} \\
 & cs, Limited (L5) & \Cor{.195} & \Cor{.374} & \Cor{.295} & \Cor{.195} & \Cor{.228} & \Cor{.273} & \Cor{.074} & \Cor{.226} & \Cor{.357} & \Cor{.236} & \Cor{.210} & \Cor{.223} & \Cor{.353} & \Cor{.154} \\
 & en, Not Limited & \Cor{.178} & \Cor{.375} & \Cor{.296} & \Cor{.187} & \Cor{.201} & \Cor{.270} & \Cor{.074} & \Cor{.251} & \Cor{.371} & \Cor{.260} & \Cor{.221} & \Cor{.233} & \Cor{.363} & \Cor{.159} \\
 & en, Limited (L5) & \Cor{.181} & \Cor{.368} & \Cor{.289} & \Cor{.181} & \Cor{.208} & \Cor{.269} & \Cor{.073} & \Cor{.277} & \Cor{.367} & \Cor{.251} & \Cor{.221} & \Cor{.229} & \Cor{.359} & \Cor{.159} \\
\midrule
\multirow{4}{*}{\rotatebox{90}{COMET}} 
 & wmt20-da & \Cor{.254} & \Cor{.468} & \Cor{.402} & \Cor{.346} & \Cor{.355} & \Cor{.308} & \Cor{.096} & \Cor{.276} & \Cor{.408} & \Cor{.302} & \Cor{.284} & \Cor{.296} & \Cor{.370} & \Cor{.182} \\
 & wmt20-qe-da & \Cor{.228} & \Cor{.267} & \Cor{.211} & \Cor{.463} & \Cor{.213} & \Cor{.283} & \Cor{.039} & \Cor{.276} & \Cor{.280} & \Cor{.293} & \Cor{.342} & \Cor{.318} & \Cor{.256} & \Cor{.125} \\
 & wmt22-da & \Cor{.269} & \Cor{.463} & \Cor{.394} & \Cor{.362} & \Cor{.369} & \Cor{.304} & \Cor{.101} & \Cor{.277} & \Cor{.381} & \Cor{.283} & \Cor{.278} & \Cor{.295} & \Cor{.360} & \Cor{.177} \\
 & cometinho-da & \Cor{.200} & \Cor{.382} & \Cor{.300} & \Cor{.224} & \Cor{.238} & \Cor{.295} & \Cor{.091} & \Cor{.237} & \Cor{.377} & \Cor{.253} & \Cor{.225} & \Cor{.226} & \Cor{.363} & \Cor{.162} \\
\midrule
\multirow{4}{*}{\rotatebox{90}{\textls[-30]{BLEURT}}} 
 & 20-D12 & \Cor{.192} & \Cor{.384} & \Cor{.306} & \Cor{.209} & \Cor{.233} & \Cor{.284} & \Cor{.082} & \Cor{.185} & \Cor{.363} & \Cor{.247} & \Cor{.214} & \Cor{.241} & \Cor{.361} & \Cor{.162} \\
 & base-512 & \Cor{.186} & \Cor{.393} & \Cor{.312} & \Cor{.238} & \Cor{.266} & \Cor{.299} & \Cor{.096} & \Cor{.239} & \Cor{.368} & \Cor{.252} & \Cor{.230} & \Cor{.244} & \Cor{.351} & \Cor{.164} \\
 & large-512 & \Cor{.185} & \Cor{.392} & \Cor{.303} & \Cor{.211} & \Cor{.245} & \Cor{.291} & \Cor{.090} & \Cor{.229} & \Cor{.358} & \Cor{.228} & \Cor{.202} & \Cor{.223} & \Cor{.355} & \Cor{.160} \\
 & tiny-128 & \Cor{.179} & \Cor{.390} & \Cor{.302} & \Cor{.207} & \Cor{.213} & \Cor{.284} & \Cor{.114} & \Cor{.215} & \Cor{.354} & \Cor{.232} & \Cor{.199} & \Cor{.206} & \Cor{.324} & \Cor{.145} \\
\bottomrule
\end{tabular}
\caption{Spearman Correlations across different metrics and model variations for MOCHA Validation and RoSE CNNDM Test (above) and WMT 21 en-ha and WMT 21 xh-zu (below).}
\label{tab:wmt21_combined_spearman}
\end{table*}

\begin{table*}[ht]
\footnotesize\centering
\setlength{\tabcolsep}{3pt}
\renewcommand{\arraystretch}{0.85}
\begin{tabular}{ll c cc cc cc @{\hspace{3mm}} c cc cc cc}
\toprule
& & \multicolumn{7}{c}{\textbf{WMT 24 cs-uk}} & \multicolumn{7}{c}{\textbf{WMT 24 en-cs}} \\
\cmidrule(lr){3-9} \cmidrule(lr){10-16}
\multirow{2}{*}{\rotatebox{90}{~~~~Metric}} & \multirow{2}{*}{Parameters} & \multirow{2}{*}{Hum} & \multicolumn{2}{c}{Llama 4} & \multicolumn{2}{c}{Llama 3.3} & \multicolumn{2}{c}{Qwen 3} & \multirow{2}{*}{GT} & \multicolumn{2}{c}{Llama 4} & \multicolumn{2}{c}{Llama 3.3} & \multicolumn{2}{c}{Qwen 3} \\
\cmidrule(lr){4-5} \cmidrule(lr){6-7} \cmidrule(lr){8-9} \cmidrule(lr){11-12} \cmidrule(lr){13-14} \cmidrule(lr){15-16}
 & & & F & Z & F & Z & F & Z & & F & Z & F & Z & F & Z \\
\midrule
\multirow{4}{*}{\rotatebox{90}{BLEU}} 
 & Order 1 & \Cor{.273} & \Cor{.469} & \Cor{.409} & \Cor{.316} & \Cor{.323} & \Cor{.423} & \Cor{.277} & \Cor{.232} & \Cor{.393} & \Cor{.339} & \Cor{.227} & \Cor{.245} & \Cor{.295} & \Cor{.209} \\
 & Order 2 & \Cor{.223} & \Cor{.434} & \Cor{.383} & \Cor{.305} & \Cor{.311} & \Cor{.421} & \Cor{.272} & \Cor{.246} & \Cor{.379} & \Cor{.324} & \Cor{.205} & \Cor{.223} & \Cor{.291} & \Cor{.202} \\
 & Order 3 & \Cor{.174} & \Cor{.415} & \Cor{.362} & \Cor{.302} & \Cor{.307} & \Cor{.419} & \Cor{.260} & \Cor{.250} & \Cor{.367} & \Cor{.312} & \Cor{.192} & \Cor{.209} & \Cor{.290} & \Cor{.198} \\
 & Order 4 & \Cor{.151} & \Cor{.394} & \Cor{.342} & \Cor{.294} & \Cor{.297} & \Cor{.416} & \Cor{.248} & \Cor{.253} & \Cor{.354} & \Cor{.299} & \Cor{.177} & \Cor{.195} & \Cor{.290} & \Cor{.194} \\
\midrule
\multirow{4}{*}{\rotatebox{90}{chrF}} 
 & c: 4, w: 0 & \Cor{.323} & \Cor{.517} & \Cor{.459} & \Cor{.333} & \Cor{.318} & \Cor{.481} & \Cor{.341} & \Cor{.229} & \Cor{.417} & \Cor{.357} & \Cor{.227} & \Cor{.229} & \Cor{.333} & \Cor{.242} \\
 & c: 4, w: 2 & \Cor{.316} & \Cor{.494} & \Cor{.433} & \Cor{.307} & \Cor{.304} & \Cor{.446} & \Cor{.310} & \Cor{.232} & \Cor{.405} & \Cor{.347} & \Cor{.214} & \Cor{.221} & \Cor{.319} & \Cor{.229} \\
 & c: 6, w: 0 & \Cor{.318} & \Cor{.508} & \Cor{.453} & \Cor{.323} & \Cor{.313} & \Cor{.474} & \Cor{.333} & \Cor{.233} & \Cor{.407} & \Cor{.350} & \Cor{.217} & \Cor{.222} & \Cor{.327} & \Cor{.235} \\
 & c: 6, w: 2 & \Cor{.315} & \Cor{.495} & \Cor{.436} & \Cor{.307} & \Cor{.306} & \Cor{.450} & \Cor{.313} & \Cor{.234} & \Cor{.402} & \Cor{.345} & \Cor{.211} & \Cor{.219} & \Cor{.320} & \Cor{.228} \\
\midrule
\multirow{4}{*}{\rotatebox{90}{ROUGE}} 
 & ROUGE-1 & \Cor{.020} & \Cor{.136} & \Cor{.120} & \Cor{.100} & \Cor{.091} & \Cor{.110} & \Cor{.066} & \Cor{.222} & \Cor{.416} & \Cor{.362} & \Cor{.245} & \Cor{.248} & \Cor{.332} & \Cor{.251} \\
 & ROUGE-2 & \Cor{.003} & \Cor{.095} & \Cor{.085} & \Cor{.067} & \Cor{.064} & \Cor{.079} & \Cor{.056} & \Cor{.211} & \Cor{.376} & \Cor{.323} & \Cor{.203} & \Cor{.206} & \Cor{.314} & \Cor{.228} \\
 & ROUGE-4 & \Cor{-.004} & \Cor{.067} & \Cor{.055} & \Cor{.034} & \Cor{.039} & \Cor{.050} & \Cor{.039} & \Cor{.122} & \Cor{.319} & \Cor{.273} & \Cor{.154} & \Cor{.154} & \Cor{.296} & \Cor{.212} \\
 & ROUGE-L & \Cor{.020} & \Cor{.136} & \Cor{.120} & \Cor{.101} & \Cor{.092} & \Cor{.110} & \Cor{.067} & \Cor{.235} & \Cor{.413} & \Cor{.363} & \Cor{.237} & \Cor{.264} & \Cor{.327} & \Cor{.252} \\
\midrule
\multirow{4}{*}{\rotatebox{90}{\textls[-40]{METEOR}}} 
 & $\alpha$: 0.2, $\gamma$: 0.0 & \Cor{.284} & \Cor{.414} & \Cor{.369} & \Cor{.274} & \Cor{.263} & \Cor{.380} & \Cor{.271} & \Cor{.214} & \Cor{.399} & \Cor{.357} & \Cor{.262} & \Cor{.270} & \Cor{.319} & \Cor{.241} \\
 & $\alpha$: 0.2, $\gamma$: 0.5 & \Cor{.236} & \Cor{.399} & \Cor{.349} & \Cor{.266} & \Cor{.261} & \Cor{.353} & \Cor{.242} & \Cor{.215} & \Cor{.388} & \Cor{.346} & \Cor{.247} & \Cor{.258} & \Cor{.307} & \Cor{.227} \\
 & $\alpha$: 0.9, $\gamma$: 0.0 & \Cor{.291} & \Cor{.423} & \Cor{.349} & \Cor{.228} & \Cor{.237} & \Cor{.360} & \Cor{.240} & \Cor{.230} & \Cor{.407} & \Cor{.343} & \Cor{.213} & \Cor{.225} & \Cor{.313} & \Cor{.224} \\
 & $\alpha$: 0.9, $\gamma$: 0.5 & \Cor{.242} & \Cor{.406} & \Cor{.338} & \Cor{.238} & \Cor{.246} & \Cor{.340} & \Cor{.222} & \Cor{.226} & \Cor{.393} & \Cor{.332} & \Cor{.203} & \Cor{.219} & \Cor{.303} & \Cor{.212} \\
\midrule
\multirow{4}{*}{\rotatebox{90}{BERTSc.}} 
 & cs, Not Limited & \Cor{.330} & \Cor{.547} & \Cor{.485} & \Cor{.372} & \Cor{.369} & \Cor{.484} & \Cor{.349} & \Cor{.236} & \Cor{.417} & \Cor{.370} & \Cor{.246} & \Cor{.255} & \Cor{.327} & \Cor{.238} \\
 & cs, Limited (L5) & \Cor{.326} & \Cor{.545} & \Cor{.487} & \Cor{.387} & \Cor{.402} & \Cor{.475} & \Cor{.337} & \Cor{.260} & \Cor{.405} & \Cor{.359} & \Cor{.239} & \Cor{.263} & \Cor{.310} & \Cor{.222} \\
 & en, Not Limited & \Cor{.311} & \Cor{.514} & \Cor{.476} & \Cor{.357} & \Cor{.352} & \Cor{.477} & \Cor{.339} & \Cor{.200} & \Cor{.415} & \Cor{.364} & \Cor{.250} & \Cor{.252} & \Cor{.325} & \Cor{.242} \\
 & en, Limited (L5) & \Cor{.321} & \Cor{.499} & \Cor{.464} & \Cor{.349} & \Cor{.343} & \Cor{.472} & \Cor{.336} & \Cor{.220} & \Cor{.409} & \Cor{.354} & \Cor{.232} & \Cor{.245} & \Cor{.311} & \Cor{.227} \\
\midrule
\multirow{4}{*}{\rotatebox{90}{COMET}} 
 & wmt20-da & \Cor{.368} & \Cor{.632} & \Cor{.596} & \Cor{.536} & \Cor{.510} & \Cor{.583} & \Cor{.437} & \Cor{.350} & \Cor{.570} & \Cor{.536} & \Cor{.474} & \Cor{.466} & \Cor{.479} & \Cor{.397} \\
 & wmt20-qe-da & \Cor{.263} & \Cor{.293} & \Cor{.339} & \Cor{.295} & \Cor{.345} & \Cor{.317} & \Cor{.177} & \Cor{.333} & \Cor{.355} & \Cor{.379} & \Cor{.430} & \Cor{.417} & \Cor{.378} & \Cor{.248} \\
 & wmt22-da & \Cor{.388} & \Cor{.619} & \Cor{.589} & \Cor{.539} & \Cor{.518} & \Cor{.573} & \Cor{.428} & \Cor{.392} & \Cor{.565} & \Cor{.536} & \Cor{.499} & \Cor{.498} & \Cor{.484} & \Cor{.400} \\
 & cometinho-da & \Cor{.361} & \Cor{.603} & \Cor{.560} & \Cor{.481} & \Cor{.459} & \Cor{.555} & \Cor{.401} & \Cor{.351} & \Cor{.513} & \Cor{.476} & \Cor{.387} & \Cor{.389} & \Cor{.412} & \Cor{.329} \\
\midrule
\multirow{4}{*}{\rotatebox{90}{\textls[-30]{BLEURT}}} 
 & 20-D12 & \Cor{.352} & \Cor{.597} & \Cor{.563} & \Cor{.494} & \Cor{.455} & \Cor{.556} & \Cor{.417} & \Cor{.375} & \Cor{.496} & \Cor{.457} & \Cor{.364} & \Cor{.377} & \Cor{.418} & \Cor{.332} \\
 & base-512 & \Cor{.292} & \Cor{.498} & \Cor{.462} & \Cor{.373} & \Cor{.346} & \Cor{.455} & \Cor{.312} & \Cor{.250} & \Cor{.408} & \Cor{.373} & \Cor{.281} & \Cor{.287} & \Cor{.324} & \Cor{.244} \\
 & large-512 & \Cor{.281} & \Cor{.485} & \Cor{.451} & \Cor{.356} & \Cor{.365} & \Cor{.446} & \Cor{.294} & \Cor{.238} & \Cor{.406} & \Cor{.367} & \Cor{.258} & \Cor{.277} & \Cor{.312} & \Cor{.228} \\
 & tiny-128 & \Cor{.221} & \Cor{.445} & \Cor{.413} & \Cor{.327} & \Cor{.339} & \Cor{.412} & \Cor{.216} & \Cor{.221} & \Cor{.443} & \Cor{.404} & \Cor{.279} & \Cor{.283} & \Cor{.351} & \Cor{.275} \\
\bottomrule
\end{tabular}

\vspace{5pt}

\begin{tabular}{ll c cc cc cc}
\toprule
& & \multicolumn{7}{c}{\textbf{WMT 24 en-is}} \\ \cmidrule(lr){3-9}
\multirow{2}{*}{\rotatebox{90}{~~~~Metric}} & & & \multicolumn{2}{c}{Llama 4} & \multicolumn{2}{c}{Llama 3.3} & \multicolumn{2}{c}{Qwen 3} \\
\cmidrule(lr){4-5} \cmidrule(lr){6-7} \cmidrule(lr){8-9}
 & Parameters & Hum & F & Z & F & Z & F & Z \\
\midrule
\multirow{4}{*}{\rotatebox{90}{BLEU}} 
 & Order 1 & \Cor{.337} & \Cor{.393} & \Cor{.318} & \Cor{.237} & \Cor{.245} & \Cor{.301} & \Cor{.201} \\
 & Order 2 & \Cor{.368} & \Cor{.387} & \Cor{.314} & \Cor{.220} & \Cor{.228} & \Cor{.299} & \Cor{.196} \\
 & Order 3 & \Cor{.372} & \Cor{.378} & \Cor{.308} & \Cor{.208} & \Cor{.214} & \Cor{.299} & \Cor{.194} \\
 & Order 4 & \Cor{.373} & \Cor{.367} & \Cor{.299} & \Cor{.192} & \Cor{.198} & \Cor{.300} & \Cor{.192} \\
\midrule
\multirow{4}{*}{\rotatebox{90}{chrF}} 
 & c: 4, w: 0 & \Cor{.319} & \Cor{.419} & \Cor{.343} & \Cor{.231} & \Cor{.222} & \Cor{.337} & \Cor{.235} \\
 & c: 4, w: 2 & \Cor{.335} & \Cor{.409} & \Cor{.333} & \Cor{.221} & \Cor{.215} & \Cor{.318} & \Cor{.219} \\
 & c: 6, w: 0 & \Cor{.333} & \Cor{.413} & \Cor{.339} & \Cor{.225} & \Cor{.218} & \Cor{.331} & \Cor{.229} \\
 & c: 6, w: 2 & \Cor{.341} & \Cor{.407} & \Cor{.333} & \Cor{.219} & \Cor{.214} & \Cor{.320} & \Cor{.220} \\
\midrule
\multirow{4}{*}{\rotatebox{90}{ROUGE}} 
 & ROUGE-1 & \Cor{.324} & \Cor{.415} & \Cor{.347} & \Cor{.244} & \Cor{.242} & \Cor{.326} & \Cor{.232} \\
 & ROUGE-2 & \Cor{.291} & \Cor{.395} & \Cor{.326} & \Cor{.224} & \Cor{.218} & \Cor{.315} & \Cor{.220} \\
 & ROUGE-4 & \Cor{.146} & \Cor{.342} & \Cor{.279} & \Cor{.172} & \Cor{.163} & \Cor{.299} & \Cor{.203} \\
 & ROUGE-L & \Cor{.347} & \Cor{.420} & \Cor{.358} & \Cor{.255} & \Cor{.273} & \Cor{.328} & \Cor{.236} \\
\midrule
\multirow{4}{*}{\rotatebox{90}{\textls[-40]{METEOR}}} 
 & $\alpha$: 0.2, $\gamma$: 0.0 & \Cor{.335} & \Cor{.406} & \Cor{.349} & \Cor{.268} & \Cor{.269} & \Cor{.314} & \Cor{.221} \\
 & $\alpha$: 0.2, $\gamma$: 0.5 & \Cor{.330} & \Cor{.398} & \Cor{.341} & \Cor{.258} & \Cor{.261} & \Cor{.309} & \Cor{.211} \\
 & $\alpha$: 0.9, $\gamma$: 0.0 & \Cor{.331} & \Cor{.397} & \Cor{.312} & \Cor{.204} & \Cor{.203} & \Cor{.299} & \Cor{.200} \\
 & $\alpha$: 0.9, $\gamma$: 0.5 & \Cor{.329} & \Cor{.389} & \Cor{.308} & \Cor{.204} & \Cor{.207} & \Cor{.296} & \Cor{.193} \\
\midrule
\multirow{4}{*}{\rotatebox{90}{BERTSc.}} 
 & cs, Not Limited & \Cor{.359} & \Cor{.425} & \Cor{.352} & \Cor{.246} & \Cor{.248} & \Cor{.325} & \Cor{.229} \\
 & cs, Limited (L5) & \Cor{.382} & \Cor{.413} & \Cor{.343} & \Cor{.243} & \Cor{.258} & \Cor{.317} & \Cor{.217} \\
 & en, Not Limited & \Cor{.308} & \Cor{.427} & \Cor{.358} & \Cor{.250} & \Cor{.248} & \Cor{.340} & \Cor{.243} \\
 & en, Limited (L5) & \Cor{.370} & \Cor{.423} & \Cor{.353} & \Cor{.241} & \Cor{.252} & \Cor{.328} & \Cor{.232} \\
\midrule
\multirow{4}{*}{\rotatebox{90}{COMET}} 
 & wmt20-da & \Cor{.443} & \Cor{.544} & \Cor{.509} & \Cor{.424} & \Cor{.416} & \Cor{.432} & \Cor{.337} \\
 & wmt20-qe-da & \Cor{.379} & \Cor{.389} & \Cor{.369} & \Cor{.405} & \Cor{.386} & \Cor{.368} & \Cor{.239} \\
 & wmt22-da & \Cor{.474} & \Cor{.535} & \Cor{.500} & \Cor{.429} & \Cor{.421} & \Cor{.413} & \Cor{.312} \\
 & cometinho-da & \Cor{.418} & \Cor{.482} & \Cor{.428} & \Cor{.349} & \Cor{.346} & \Cor{.385} & \Cor{.288} \\
\midrule
\multirow{4}{*}{\rotatebox{90}{\textls[-30]{BLEURT}}} 
 & 20-D12 & \Cor{.416} & \Cor{.450} & \Cor{.394} & \Cor{.307} & \Cor{.313} & \Cor{.367} & \Cor{.265} \\
 & base-512 & \Cor{.335} & \Cor{.422} & \Cor{.359} & \Cor{.289} & \Cor{.291} & \Cor{.338} & \Cor{.243} \\
 & large-512 & \Cor{.342} & \Cor{.407} & \Cor{.337} & \Cor{.240} & \Cor{.268} & \Cor{.325} & \Cor{.221} \\
 & tiny-128 & \Cor{.331} & \Cor{.442} & \Cor{.380} & \Cor{.278} & \Cor{.269} & \Cor{.353} & \Cor{.271} \\
\bottomrule
\end{tabular}
\caption{Spearman Correlations across different metrics and model variations for WMT 24 cs-uk and WMT 24 en-cs (above) and WMT 24 en-is (below).}
\label{tab:wmt24_enis_spearman}
\end{table*}


\begin{table*}[t]
\begin{lstlisting}
You are a Semantic Corruption Engine for NLP evaluation.
Your task is to generate a single "synthetic hypothesis" string by modifying the provided 'input_answer' based on the requested 'damage_level'.

### GROUND TRUTH PROTOCOL
1. **Facts:** Treat the 'input_answer' as the absolute factual truth for this task.
   - At Level 0, you must agree with the 'input_answer'.
   - At Level 5, you must contradict the 'input_answer'.
2. **Context:** Use the provided 'question' to understand the topic, gender, and grammatical context required for the answer.

### DAMAGE SPECIFICATIONS
Level 0 (Paraphrase): Rewrite the 'input_answer' using different words or grammar, but strictly preserve the original meaning.
Level 1 (Surface Noise): Keep the meaning true. You may remove minor adjectives, generalize numbers, or simplify phrasing.
Level 2 (Omission): Remove a specific detail (like a name, date, or location). Make the answer vaguely true but less informative.
Level 3 (Minor Semantic Error): Keep the topic, but alter a specific entity to a plausible but incorrect one (e.g., swap a city for a nearby town, change a date by a few years).
Level 4 (Major Semantic Error): Significantly alter the meaning. Swap the main Subject or Object to something clearly wrong but related (e.g., change the actor to a different actor).
Level 5 (Hallucination): Produce a fluent, confident answer to the 'question' that is completely factually wrong compared to the 'input_answer'.

### CONSTRAINTS
1. OUTPUT LANGUAGE: The output must be in the SAME LANGUAGE and script as the 'input_answer' (e.g., if input is Czech, output must be Czech).
2. FORMAT: Output ONLY the resulting text string. Do not include labels like "Output:" or explanations.
\end{lstlisting}

\caption{Prompt for zero-shot synthetic data generation for question answering on the CUS-QA dataset.}\label{tab:prompt_cusqa_zero}
\end{table*}

\begin{table*}
\begin{lstlisting}
You are a Semantic Corruption Engine for Reading Comprehension.
Your task is to generate a single "synthetic text" string by modifying the provided 'input_answer' based on the requested 'damage_level'.

### GROUND TRUTH PROTOCOL
1. **Source of Truth:** The 'passage' is the absolute factual truth. Any deviation from the passage counts as damage.
2. **Relevance:** The output must still attempt to answer the 'question', even if the facts are modified (at higher levels).

### DAMAGE SPECIFICATIONS
Level 0 (Paraphrase): Rewrite the 'input_answer' using different words or syntax. You MUST preserve the exact meaning supported by the 'passage'.
Level 1 (Surface Noise): Keep the meaning true. You may remove minor adjectives, generalize numbers slightly, or simplify phrasing.
Level 2 (Loss of Precision): Omit a secondary detail or make the answer slightly less specific than the 'input_answer'.
Level 3 (Minor Semantic Error): Alter a specific entity or relationship. Swap a name, date, or location with a plausible but incorrect one not supported by the 'passage'.
Level 4 (Major Semantic Error): Significantly alter the core meaning. Swap the main Subject/Object or negate the main verb.
Level 5 (Total Hallucination): Produce a fluent, confident answer that is completely unsupported by the 'passage' or explicitly contradicts it.

### CONSTRAINTS
1. OUTPUT LANGUAGE: English (unless the input is in another language).
2. FORMAT: Output ONLY the resulting text string. Do not include labels, explanations, or quotes.
\end{lstlisting}

\caption{Prompt for zero-shot synthetic data generation for question answering on the MOCHA dataset.}\label{tab:prompt_mocha_zero}
\end{table*}

\begin{table*}
\begin{lstlisting}
You are a Semantic Corruption Engine for NLP evaluation.
Your task is to generate a single "synthetic hypothesis" string by modifying the provided 'input_answer' based on the requested 'damage_level'.

### GROUND TRUTH PROTOCOL
1. **Facts:** Treat the 'input_answer' as the absolute factual truth for this task.
   - At Level 0, you must agree with the 'input_answer'.
   - At Level 5, you must contradict the 'input_answer'.
2. **Context:** Use the provided 'question' to understand the topic, gender, and grammatical context required for the answer.

### DAMAGE SPECIFICATIONS
Level 0 (Paraphrase): Rewrite the 'input_answer' using different words or grammar, but strictly preserve the original meaning.
Level 1 (Surface Noise): Keep the meaning true. You may remove minor adjectives, generalize numbers, or simplify phrasing.
Level 2 (Omission): Remove a specific detail (like a name, date, or location). Make the answer vaguely true but less informative.
Level 3 (Minor Semantic Error): Keep the topic, but alter a specific entity to a plausible but incorrect one (e.g., swap a city for a nearby town, change a date by a few years).
Level 4 (Major Semantic Error): Significantly alter the meaning. Swap the main Subject or Object to something clearly wrong but related (e.g., change the actor to a different actor).
Level 5 (Hallucination): Produce a fluent, confident answer to the 'question' that is completely factually wrong compared to the 'input_answer'.

### EXAMPLES

User:
question: Who directed the movie 'Titanic'?
input_answer: James Cameron directed it.
damage_level: 0

Assistant:
The film was directed by James Cameron.

User:
question: What is the capital of Slovakia?
input_answer: Bratislava.
damage_level: 3

Assistant:
Košice.

User:
question: Jaké je hlavní město České republiky?
input_answer: Hlavním městem je Praha.
damage_level: 5

Assistant:
Hlavním městem je Ostrava, známá svými plážemi.

### CONSTRAINTS
1. OUTPUT LANGUAGE: The output must be in the SAME LANGUAGE and script as the 'input_answer'.
2. FORMAT: Output ONLY the resulting text string. Do not include labels like "Output:" or explanations.
\end{lstlisting}

\caption{Prompt for few-shot synthetic data generation for question answering on the CUS-QA dataset.}\label{tab:prompt_cusqa_few}
\end{table*}

\begin{table*}
\begin{lstlisting}
You are a Semantic Corruption Engine for Reading Comprehension.
Your task is to generate a single "synthetic text" string by modifying the provided 'input_answer' based on the requested 'damage_level'.

### GROUND TRUTH PROTOCOL
1. **Source of Truth:** The 'passage' is the absolute factual truth. Any deviation from the passage counts as damage.
2. **Relevance:** The output must still attempt to answer the 'question', even if the facts are modified (at higher levels).

### DAMAGE SPECIFICATIONS
Level 0 (Paraphrase): Rewrite the 'input_answer' using different words or syntax. You MUST preserve the exact meaning supported by the 'passage'.
Level 1 (Surface Noise): Keep the meaning true. You may remove minor adjectives, generalize numbers slightly, or simplify phrasing.
Level 2 (Loss of Precision): Omit a secondary detail or make the answer slightly less specific than the 'input_answer'.
Level 3 (Minor Semantic Error): Alter a specific entity or relationship. Swap a name, date, or location with a plausible but incorrect one not supported by the 'passage'.
Level 4 (Major Semantic Error): Significantly alter the core meaning. Swap the main Subject/Object or negate the main verb.
Level 5 (Total Hallucination): Produce a fluent, confident answer that is completely unsupported by the 'passage' or explicitly contradicts it.

### CONSTRAINTS
1. OUTPUT LANGUAGE: English (unless the input is in another language).
2. FORMAT: Output ONLY the resulting text string. Do not include labels, explanations, or quotes.

### EXAMPLES

User:
passage: The Apollo 11 mission landed humans on the Moon in July 1969.
question: When did the landing occur?
input_answer: It happened in 1969.
damage_level: 0

Assistant:
The landing took place in the year 1969.

User:
passage: Photosynthesis takes place inside the chloroplasts, which contain chlorophyll.
question: Where does photosynthesis happen?
input_answer: It occurs in the chloroplasts.
damage_level: 3

Assistant:
It occurs in the mitochondria.

User:
passage: The blue whale is the largest animal known to have ever lived.
question: What is the largest animal?
input_answer: The blue whale.
damage_level: 5

Assistant:
The largest animal is the African Elephant.
\end{lstlisting}

\caption{Prompt for few-shot synthetic data generation for question answering on the MOCHA dataset.}\label{tab:prompt_mocha_few}
\end{table*}

\begin{table*}[t]
\begin{lstlisting}
You are an Atomic Fact Corruption Engine.
Your task is to generate a "synthetic text" by modifying a 'reference_summary' based on a 'damage_level', specifically targeting Atomic Content Units (ACUs).

### THE ACU PROTOCOL
Summaries are evaluated by breaking them down into "Atomic Content Units" (fine-grained, independent facts) and checking their recall.
- **Goal:** As Damage Level increases, the number of ACUs from the 'reference_summary' preserved in your output must DECREASE.
- **Constraint:** You must maintain the **fluency** and **length** of the original text. Do not simply delete sentences; replace facts with non-facts or plausible hallucinations.

### DAMAGE SPECIFICATIONS (ACU RECALL)
Level 0 (100% ACU Recall): Paraphrase the text but preserve **every single atomic fact** (names, dates, relations, quantities).
Level 1 (80% ACU Recall): Preserve the main story but blur specific details. (e.g., Change "David Ospina" to "the goalkeeper", or "16th minute" to "early on").
Level 2 (60% ACU Recall): Remove minor ACUs. Replace specific facts with generic filler text that sounds relevant but conveys no specific information from the source.
Level 3 (40% ACU Recall): Entity Swap. Keep the sentence structure but swap key entities (Subject/Object) so the ACUs become factually false (e.g., "Chelsea won" -> "Arsenal won").
Level 4 (20% ACU Recall): Major Contradiction. Rewrite the summary to describe a different outcome or event involving the same entities, falsifying nearly all original facts.
Level 5 (0% ACU Recall): Total Hallucination. Generate a fluent summary of the same length that contains **ZERO** facts from the reference. It can be about the same topic but must be factually disjoint.

### CONSTRAINTS
1. LENGTH: The output must be within ±10% word count of the 'reference_summary'.
2. FLUENCY: The text must be grammatically perfect.
3. FORMAT: Output ONLY the resulting summary string. No labels.
\end{lstlisting}
\caption{Prompt for zero-shot synthetic data generation for summarization on the RoSE dataset.}\label{tab:prompt_rose_zero}
\end{table*}

\begin{table*}
\begin{lstlisting}
You are a Translation Corruption Engine.
Your goal is to take a perfect 'reference_translation' and degrade it according to the specific 'damage_level' requested.

### GROUND TRUTH PROTOCOL
1. **Source of Truth:** The 'source_sentence' and 'reference_translation' define the correct meaning.
2. **Strict adherence:** You must NOT improve the text. You must damage it.

### DAMAGE SPECIFICATIONS
Level 0 (Paraphrase): Rewrite the 'reference_translation' using different synonyms or sentence structures. It MUST remain a valid, high-quality translation of the 'source_sentence' with perfect grammar.
Level 1 (Surface/Mechanical Noise): Keep the words mostly identical to the 'reference_translation', but **inject a visible technical error**. You MUST include a spelling mistake, a capitalization error, missing punctuation, or a blatant subject-verb agreement error (e.g., "he go" instead of "he goes"). The meaning must remain perfect, but the fluency must be damaged.
Level 2 (Omission/Under-translation): Remove a specific detail or nuance found in the 'source_sentence' (e.g., drop an adjective or adverb). The translation is understandable but clearly incomplete compared to the reference.
Level 3 (Word-Level Semantic Error): Mistranslate a specific content word (noun/verb) to a plausible but incorrect alternative (e.g., "car" -> "truck", "walked" -> "ran"). This must be a specific, local error.
Level 4 (Major Semantic Error): Significantly alter the meaning of the whole sentence. Swap the Subject and Object, negate the main verb, or change the tense dramatically (past -> future) if it contradicts the source.
Level 5 (Hallucination/Catastrophic Failure): Produce a fluent sentence in the target language that has NOTHING to do with the 'source_sentence', or is a translation of a completely different input.

### CONSTRAINTS
1. OUTPUT LANGUAGE: The output must be in the SAME LANGUAGE as the 'reference_translation'.
2. NO EXACT MATCHES: For Damage Level 1 and above, the output **MUST NOT** be identical to the 'reference_translation'.
3. FORMAT: Output ONLY the resulting translation string. No labels, no explanations.
\end{lstlisting}
\caption{Prompt for zero-shot synthetic data generation for machine translation.}\label{tab:prompt_mt_zero}
\end{table*}

\begin{table*}
\begin{lstlisting}    
You are an Atomic Fact Corruption Engine.
Your task is to generate a "synthetic text" by modifying a 'reference_summary' based on a 'damage_level', specifically targeting Atomic Content Units (ACUs).

### THE ACU PROTOCOL
Summaries are evaluated by breaking them down into "Atomic Content Units" (fine-grained, independent facts) and checking their recall.
- **Goal:** As Damage Level increases, the number of ACUs from the 'reference_summary' preserved in your output must DECREASE.
- **Constraint:** You must maintain the **fluency** and **length** of the original text. Do not simply delete sentences; replace facts with non-facts or plausible hallucinations.

### DAMAGE SPECIFICATIONS (ACU RECALL)
Level 0 (100% ACU Recall): Paraphrase the text but preserve **every single atomic fact** (names, dates, relations, quantities).
Level 1 (80% ACU Recall): Preserve the main story but blur specific details. (e.g., Change "David Ospina" to "the goalkeeper", or "16th minute" to "early on").
Level 2 (60% ACU Recall): Remove minor ACUs. Replace specific facts with generic filler text that sounds relevant but conveys no specific information from the source.
Level 3 (40% ACU Recall): Entity Swap. Keep the sentence structure but swap key entities (Subject/Object) so the ACUs become factually false (e.g., "Chelsea won" -> "Arsenal won").
Level 4 (20% ACU Recall): Major Contradiction. Rewrite the summary to describe a different outcome or event involving the same entities, falsifying nearly all original facts.
Level 5 (0% ACU Recall): Total Hallucination. Generate a fluent summary of the same length that contains **ZERO** facts from the reference. It can be about the same topic but must be factually disjoint.

### CONSTRAINTS
1. LENGTH: The output must be within ±10% word count of the 'reference_summary'.
2. FLUENCY: The text must be grammatically perfect.
3. FORMAT: Output ONLY the resulting summary string. No labels.

### EXAMPLES

User:
source_text: (Context about the Chelsea vs Arsenal match)
reference_summary: Chelsea weren't awarded a penalty for David Ospina's clash with Oscar. Arsenal goalkeeper clattered Oscar inside the box.
damage_level: 0

Assistant:
David Ospina collided with Oscar in the box, but Chelsea did not receive a penalty for the incident involving the Arsenal goalkeeper.

User:
source_text: (Context about the Chelsea vs Arsenal match)
reference_summary: Chelsea weren't awarded a penalty for David Ospina's clash with Oscar. Arsenal goalkeeper clattered Oscar inside the box.
damage_level: 3

Assistant:
Arsenal weren't awarded a penalty for Petr Cech's clash with Willian. The Chelsea goalkeeper clattered Willian outside the box.

User:
source_text: (Context about SpaceX Starship)
reference_summary: SpaceX's Starship rocket exploded four minutes after launch on Thursday, with no injuries reported.
damage_level: 5

Assistant:
NASA's Falcon 9 booster landed successfully on the drone ship on Friday, marking another milestone for the agency.
\end{lstlisting}

\caption{Prompt for few-shot synthetic data generation for summarization on the RoSE dataset.}\label{tab:prompt_rose_few}
\end{table*}

\begin{table*}
\begin{lstlisting}
You are a Translation Corruption Engine.
Your goal is to take a perfect 'reference_translation' and degrade it according to the specific 'damage_level' requested.

### GROUND TRUTH PROTOCOL
1. **Source of Truth:** The 'source_sentence' and 'reference_translation' define the correct meaning.
2. **Strict adherence:** You must NOT improve the text. You must damage it.

### DAMAGE SPECIFICATIONS
Level 0 (Paraphrase): Rewrite the 'reference_translation' using different synonyms or sentence structures. It MUST remain a valid, high-quality translation of the 'source_sentence' with perfect grammar.
Level 1 (Surface/Mechanical Noise): Keep the words mostly identical to the 'reference_translation', but **inject a visible technical error**. You MUST include a spelling mistake, a capitalization error, missing punctuation, or a blatant subject-verb agreement error (e.g., "he go" instead of "he goes"). The meaning must remain perfect, but the fluency must be damaged.
Level 2 (Omission/Under-translation): Remove a specific detail or nuance found in the 'source_sentence' (e.g., drop an adjective or adverb). The translation is understandable but clearly incomplete compared to the reference.
Level 3 (Word-Level Semantic Error): Mistranslate a specific content word (noun/verb) to a plausible but incorrect alternative (e.g., "car" -> "truck", "walked" -> "ran"). This must be a specific, local error.
Level 4 (Major Semantic Error): Significantly alter the meaning of the whole sentence. Swap the Subject and Object, negate the main verb, or change the tense dramatically (past -> future) if it contradicts the source.
Level 5 (Hallucination/Catastrophic Failure): Produce a fluent sentence in the target language that has NOTHING to do with the 'source_sentence', or is a translation of a completely different input.

### CONSTRAINTS
1. OUTPUT LANGUAGE: The output must be in the SAME LANGUAGE as the 'reference_translation'.
2. NO EXACT MATCHES: For Damage Level 1 and above, the output **MUST NOT** be identical to the 'reference_translation'.
3. FORMAT: Output ONLY the resulting translation string. No labels, no explanations.

### EXAMPLES

User:
source_sentence: The cat sat on the mat.
reference_translation: Le chat s'est assis sur le tapis.
damage_level: 0

Assistant:
Le chat était assis sur le tapis.

User:
source_sentence: She bought a red car yesterday.
reference_translation: Sie hat gestern ein rotes Auto gekauft.
damage_level: 3

Assistant:
Sie hat gestern ein blaues Fahrrad gekauft.

User:
source_sentence: Technology is evolving rapidly.
reference_translation: La technologie évolue rapidement.
damage_level: 5

Assistant:
J'aime manger des pommes au petit déjeuner.
\end{lstlisting}

\caption{Prompt for few-shot synthetic data generation for machine translation.}\label{tab:prompt_mt_few}
\end{table*}

\begin{table*}[t]
\begin{lstlisting}
### DAMAGE SPECIFICATIONS
Level 0 (Semantic Equivalence): Reformulate the input text using novel
  vocabulary or syntax while retaining 100% of the original information
  and meaning.
Level 1 (Lossy Compression): Maintain factual accuracy but degrade
  precision. Approximate numerical values, strip non-essential
  modifiers, or use simpler diction.
Level 2 (Information Redaction): Omit key identifiers (such as specific
  names, dates, or locations). The statement should remain truthful but
  lack specificity.
Level 3 (Low-Level Distortion): Retain the general context but
  substitute a single specific entity with a plausible but incorrect
  alternative (e.g., shifting a date slightly or swapping a city for
  a neighbor).
Level 4 (High-Level Distortion): Fundamentally invalidate the core
  meaning. Replace the primary Subject or Object with an entity that
  is contextually related but demonstrably incorrect.
Level 5 (Complete Fabrication): Generate a hallucinated response that
  answers the question with high confidence and fluency, but is
  diametrically opposed to the facts in the input answer.
\end{lstlisting}
\caption{Damage specifications for CUS-QA sensitivity Variant~1. The surrounding system role, ground-truth protocol, and output constraints are identical to Table~\ref{tab:prompt_cusqa_zero}.}\label{tab:prompt_sensitivity_v1}
\end{table*}

\begin{table*}[t]
\begin{lstlisting}
### DAMAGE SPECIFICATIONS
Level 0 (Restate): Say exactly the same thing as the input answer,
  but say it differently. Do not change the facts.
Level 1 (Simplify): Keep the answer correct, but make it rougher.
  Remove specific adjectives or round off numbers to be less precise.
Level 2 (Delete Details): Remove specific proper nouns or data points.
  The sentence should still be technically true, just vague.
Level 3 (Tweak Entity): Keep the general topic correct, but swap one
  specific detail for a close semantic neighbor
  (e.g., Tuesday -> Wednesday, Paris -> Lyon).
Level 4 (Break Relationship): Change the main actor or target of the
  sentence. Make the statement clearly wrong by swapping the key
  subject/object.
Level 5 (Lie): Ignore the truth. Write a smooth, well-phrased answer
  to the question that is completely false compared to the input answer.
\end{lstlisting}
\caption{Damage specifications for CUS-QA sensitivity Variant~2 (colloquial action-label style).}\label{tab:prompt_sensitivity_v2}
\end{table*}

\begin{table*}[t]
\begin{lstlisting}
### DAMAGE SPECIFICATIONS
Level 0 (True - Rephrased): The output must be factually identical to
  the input, differing only in style or phrasing.
Level 1 (True - Noisy): The output remains factually compatible with
  the input, but loses fine-grained detail or precision.
Level 2 (True - Incomplete): The output is factually compatible but
  insufficient; critical specifics like names or dates are missing.
Level 3 (False - Plausible): The output contains a small factual
  error. It sounds correct but includes a near-miss entity
  substitution.
Level 4 (False - Major): The output contains a significant factual
  error regarding the main subject or object, rendering the core
  statement false.
Level 5 (False - Total): The output is a complete fabrication. It
  presents a confident, fictional reality that contradicts the input
  entirely.
\end{lstlisting}
\caption{Damage specifications for CUS-QA sensitivity Variant~3. The surrounding system role, ground-truth protocol, and output constraints are identical to Table~\ref{tab:prompt_cusqa_zero}.}\label{tab:prompt_sensitivity_v3}
\end{table*}

\begin{table*}[t]
\begin{lstlisting}
### DAMAGE SPECIFICATIONS
Level 0 (Synonym Swap): Edit the sentence structure and vocabulary
  without altering the underlying logic or facts.
Level 1 (Generalization): Edit the text to be less specific. Convert
  exact figures to ranges or estimates; remove descriptive flair.
Level 2 (Redaction): Edit the text to remove proper nouns
  (Who/Where/When). Leave the "What" intact but vague.
Level 3 (Minor Glitch): Edit a single entity. Change a specific detail
  to something that looks similar but is factually wrong.
Level 4 (Major Swap): Edit the key players. Change the Subject or
  Object to a different entity, breaking the factual link.
Level 5 (Creative Writing): Discard the facts. Write a convincing but
  entirely invented answer to the prompt.
\end{lstlisting}
\caption{Damage specifications for CUS-QA sensitivity Variant~4. The surrounding system role, ground-truth protocol, and output constraints are identical to Table~\ref{tab:prompt_cusqa_zero}.}\label{tab:prompt_sensitivity_v4}
\end{table*}

\begin{table*}[t]
\begin{lstlisting}
### DAMAGE SPECIFICATIONS
Level 0: Paraphrase. Keep meaning exact.
Level 1: Generalize. Remove adjectives, simplify numbers.
         Keep meaning true.
Level 2: Omit. Remove names/dates/locations. Answer becomes vague.
Level 3: Minor Error. Swap one entity for a plausible incorrect one.
Level 4: Major Error. Swap the main Subject or Object.
         Meaning is now false.
Level 5: Hallucinate. Generate a confident, fluent, but totally
         false answer.
\end{lstlisting}
\caption{Damage specifications for CUS-QA sensitivity Variant~5. The surrounding system role, ground-truth protocol, and output constraints are identical to Table~\ref{tab:prompt_cusqa_zero}.}\label{tab:prompt_sensitivity_v5}
\end{table*}

\end{document}